\documentclass{article}

\PassOptionsToPackage{numbers, compress}{natbib}

\usepackage[preprint]{neurips_2026}

\usepackage[utf8]{inputenc} 
\usepackage[T1]{fontenc}    
\usepackage{hyperref}       
\usepackage{url}            
\usepackage{microtype}
\usepackage{paralist} 
\usepackage{graphicx}
\usepackage{subcaption}
\usepackage{booktabs}       
\usepackage{alltt}
\usepackage{caption}
\usepackage{tabularx}
\usepackage{enumitem}
\usepackage{paralist}
\usepackage[table]{xcolor}
\usepackage{multirow}
\usepackage{tcolorbox}
\usepackage[numbers]{natbib}
\usepackage{algorithm}
\usepackage{algorithmic}

\usepackage{amsmath}
\usepackage{amsfonts}       
\usepackage{amssymb}
\usepackage{mathtools}
\usepackage{amsthm}
\usepackage{nicefrac}       

\usepackage[capitalize,noabbrev]{cleveref}

\theoremstyle{plain}

\theoremstyle{definition}

\theoremstyle{remark}

\usepackage[textsize=tiny]{todonotes}

\usepackage{listings}
\usepackage{xcolor}
\usepackage{float} 
\usepackage{placeins}

\title{Physical Context Based Benchmarking Metrics for Multimodal Synthetic Images}

\author{%
   Kishor Datta Gupta\thanks{Corresponding author.} \\
  \And
   Marufa Kamal\footnotemark[1] 
   \And
   Md.\ Mahfuzur Rahman 
   \And
   Fahad Rahman 
   \And
   Mohd Ariful Haque 
   \And
   Sunzida Siddique 
}

\begin{document}

\maketitle

\begin{abstract}
Current state-of-the-art evaluation metrics, including BLEU, CIDEr, VQA score, SigLIP-2, and CLIPScore, often fail to capture semantic correctness and structural consistency, particularly in domain specific or context dependent settings. To address these limitations, this paper introduces a Physical-Context-based Multimodal Data Evaluation (PCMDE) metric that integrates vision language models(VLM) with knowledge based mapping and physical context guided reasoning using LLMs. The proposed framework consists of three main stages: (1) multimodal feature extraction of spatial and semantic information using object detection and VLM; (2) confidence weighted component fusion for adaptive, component level validation; and (3) physical context guided reasoning using LLMs to enforce structural and relational constraints, like alignment, relative position, and consistency. Unlike semantic-alignment metrics, PCMDE evaluates whether generated images satisfy component-level physical constraints and returns both a calibrated plausibility score and an interpretable diagnostic message identifying the violated structural rules.
\end{abstract}

\vspace{-10pt}
\section{Introduction}

Synthetic images generated from text descriptions have become important in many real world applications, from autonomous systems to medical imaging. However, ensuring that these images are physically correct remains a significant challenge. Current evaluation metrics focus on measuring semantic alignment between generated images and text prompts but often fail to capture structural accuracy, especially in domain specific scenarios. CLIPScore~\cite{hessel2022clipscorereferencefreeevaluationmetric} measures cosine similarity between CLIP encoded image and text embeddings to assess caption alignment, but struggles with compositional reasoning, it cannot reliably evaluate spatial relationships or object counts. An aircraft image with engines positioned upside-down on the wings (Fig. 1a) receives a high CLIPScore because the metric identifies ``aircraft'', ``engines'' and ``wings'' despite violating aerodynamic principles. VQAScore~\cite{lin2024evaluatingtexttovisualgenerationimagetotext} improves upon CLIPScore by using visual question answering to verify attributes, yet it remains limited to binary assessments and cannot evaluate continuous structural integrity. A car with wheels above the roofline (Fig. 1b) may pass VQAScore's checks while violating the constraint that wheels must support the vehicle. SigLIP-2~\cite{tschannen2025siglip2multilingualvisionlanguage}, an improved CLIP variant, shows even more severe limitations in preliminary experiments (Tables 1 and Case study in Appendix), it produces near uniform scores (coefficient of variation \textless 1.5), unable to distinguish between physically correct and incorrect configurations. These metrics operate on statistical correlations without explicit reasoning about physical constraints.

\begin{figure}[t]
    \centering
    \begin{subfigure}[t]{0.3\linewidth}
        \centering
        \includegraphics[width=\linewidth]{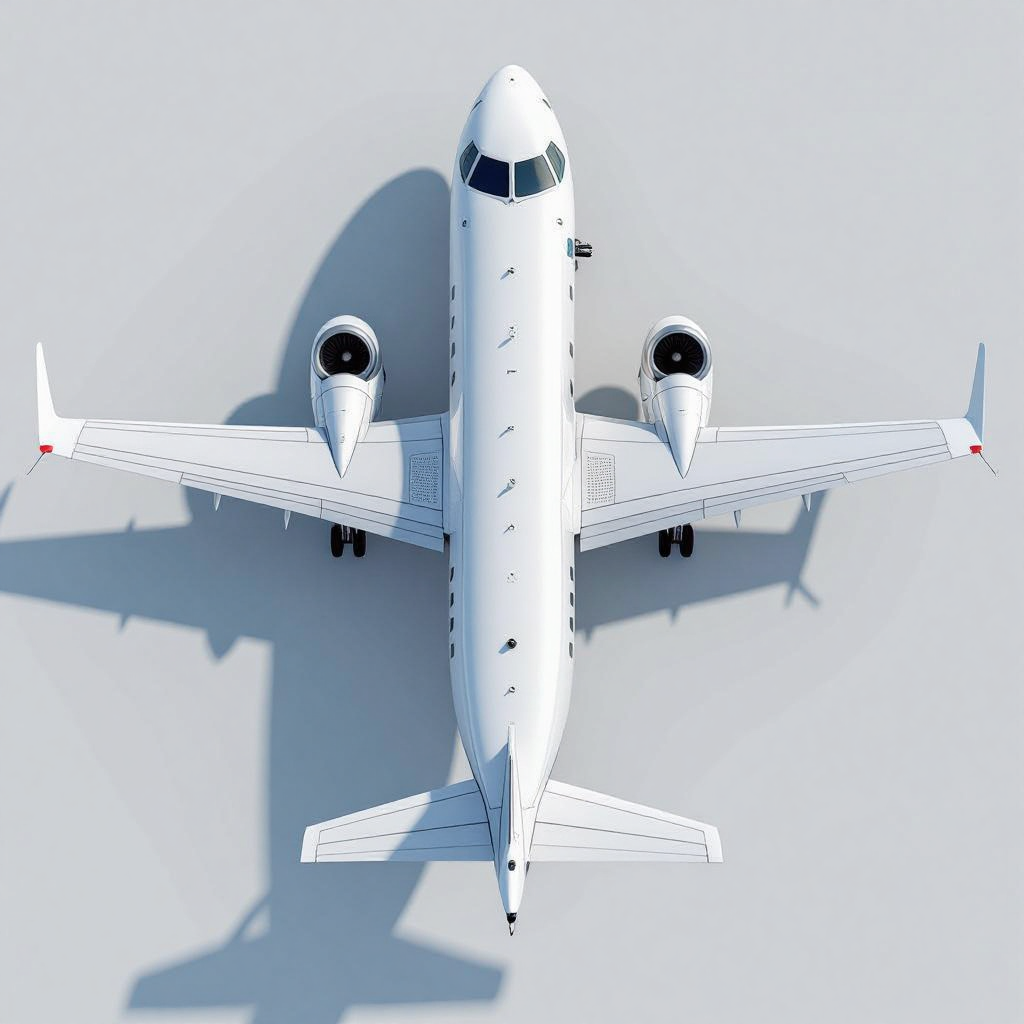}
        \caption{Aircraft image with physically impossible structure.}
        \label{fig:aircraft_misplacement}
    \end{subfigure}
    \hfill
    \begin{subfigure}[t]{0.3\linewidth}
        \centering
        \includegraphics[width=\linewidth]{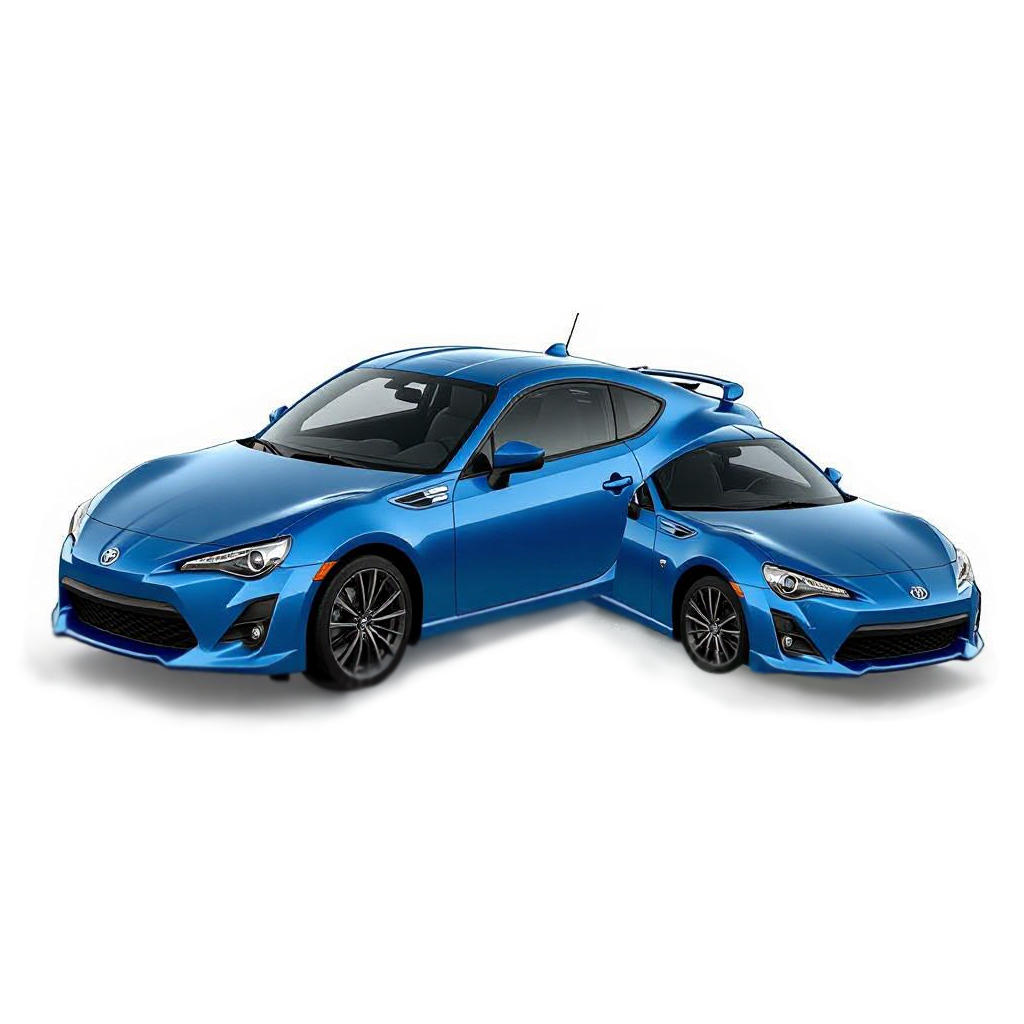}
        \caption{Car image with physically impossible structure.}
        \label{fig:car_misplacement}
    \end{subfigure}
    \hfill
    \begin{subfigure}[t]{0.3\linewidth}
        \centering
        \includegraphics[width=\linewidth]{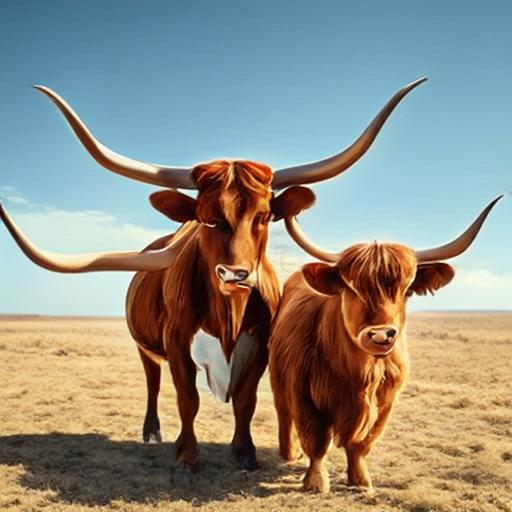}
        \caption{Cattle image with physically impossible anatomical structure.}
        \label{fig:cattle_misplacement}
    \end{subfigure}
    
    \caption{\footnotesize Visually realistic but structurally incorrect images. These examples violate fundamental aerodynamic, vehicular, or anatomical structures. Multiple metrics (CLIPScore, VQA Score, and SigLIP-2 etc.) demonstrate the similarity of image and text features. While image and text features belong to different domains, we do not directly compare image and text; a rule-based approach is used to refine the final result using VLM with LLM reasoning.}
    \label{fig:structural_misplacement}
\end{figure}

Recent work has attempted to incorporate physical reasoning. Science-T2I~\cite{li2025sciencet2iaddressingscientificillusions} created a dataset of 20,000 image pairs covering 16 phenomena across physics, chemistry and biology. PhyBench~\cite{meng2024phybenchphysicalcommonsensebenchmark} established a benchmark with 700 prompts spanning 31 physical scenarios. While both represent progress beyond semantic matching, they focus on general physical principles rather than domain specific structural validation. Neither performs component level detection; they cannot identify whether an aircraft has the correct number of engines or verify that car wheels are at the bottom of the vehicle body. This makes them insufficient for specialized domains like aerospace or automotive engineering, where correctness depends on detecting specific components and verifying their spatial arrangements according to domain rules.

We propose PCMDE (Physical Context based Multimodal Data Evaluation), a framework that evaluates synthetic images by checking components and their spatial relationships. PCMDE operates through three stages: (1) multi source component detection combining a domain specific object detector with VLMs for robust identification; (2) rule based physical validation checking presence (required components in correct quantities), spatial (components in valid positions) and relational constraints (logical dependencies between parts); and (3) LLM-based reasoning verifying that configurations match type specific constraints (e.g., DC-10 aircraft require exactly 3 engines, sedan cars require 4 doors). Unlike embedding based metrics, PCMDE provides interpretable diagnostics identifying specific violations. This work identifies limitations in existing metrics for domain specific structural assessment, proposes a framework combining component detection with physical constrained rules and LLM reasoning, and provides an interpretable framework for evaluating synthetic images where structural correctness is essential. PCMDE is intended as an evaluation artifact rather than a generative model. It specifies what is measured, component-level physical plausibility; the assumptions under which the score is meaningful, domains with detectable parts and explicit structural rules; and the limits of interpretation, including detector errors, viewpoint ambiguity, and incomplete rule coverage. Thus, PCMDE complements semantic alignment metrics by evaluating structural validity rather than replacing metrics that measure caption-image similarity.
\vspace{-10pt}
\section{Related Works}\vspace{-10pt}
Synthetic Data Generation addresses data scarcity in ML by producing annotated datasets. Early methods-procedural modeling, simulation, and domain randomization~\cite{merrell2010model, nedevschi2019semantic, Tobin} offered scalability but limited semantic control and often induced bias. Modern GenAI models (GANs, VAEs, VQAs, diffusion)~\cite{tamayo2024gan,wu2025synthetic,liu2025synthvlm} create photorealistic data but often lack physical and structural consistency~\cite{jimaging11080252}. Synthetic Data Evaluation is based on metrics such as BLEU, CIDEr and Inception~\cite{tran2019does, oliveira2021cider, chan2024evaluating} for semantic and visual realism, but they overlook contextual and relational consistency~\cite{zamzmi2025scorecard}. Embedding metrics (CLIPScore, InfoMetric)~\cite{hessel2022clipscorereferencefreeevaluationmetric} help with alignment but miss structural integrity and rule-based correctness. Multimodal Fusion and Vision-Language Models(e.g., CLIP, BLIP-2, Gemini, Qwen-VL)~\cite{han2025multimodal} integrate vision and text via cross-attention or adapters for semantic understanding and manipulation tasks. LLMs apply reasoning and domain constraints to synthesize realistic, diverse datasets~\cite{fedoseev24constraint, zebaze2025llm}, improving generalization in low-data~\cite{goldie2025synthetic}. Some evaluation matrices measure semantic similarity using contextual embeddings~\cite{zhang2019bertscore},  METEOR~\cite{banerjee2005meteor} considers synonymy, stemming, and recall in the evaluation of captions; ROUGE~\cite{schluter2017limits} evaluates the degree of overlap in phrases and sequences, while SPICE~\cite{anderson2016spice} assesses semantic scene through the object-relation graphs. These insights motivate our proposed PCMDE framework, combining generative, physics-guided, and multimodal fusion techniques for robust synthetic data benchmarking.

\vspace{-13pt}
\section{Proposed Methodology} \vspace{-10pt} 
Our proposed Physical-Context-based Multimodal Data Evaluation (PCMDE) is a framework for assessing structural realism in synthetic images by combining object detection, physically constrained rules, and LLM reasoning. The framework operates through three sequential stages: detecting image components from multiple sources, validating component arrangements against physical constraints, and applying specification-aware reasoning to capture domain-specific knowledge that cannot be easily encoded as fixed rules.

\begin{figure*}[t]
\centering
\includegraphics[width=0.9\linewidth]{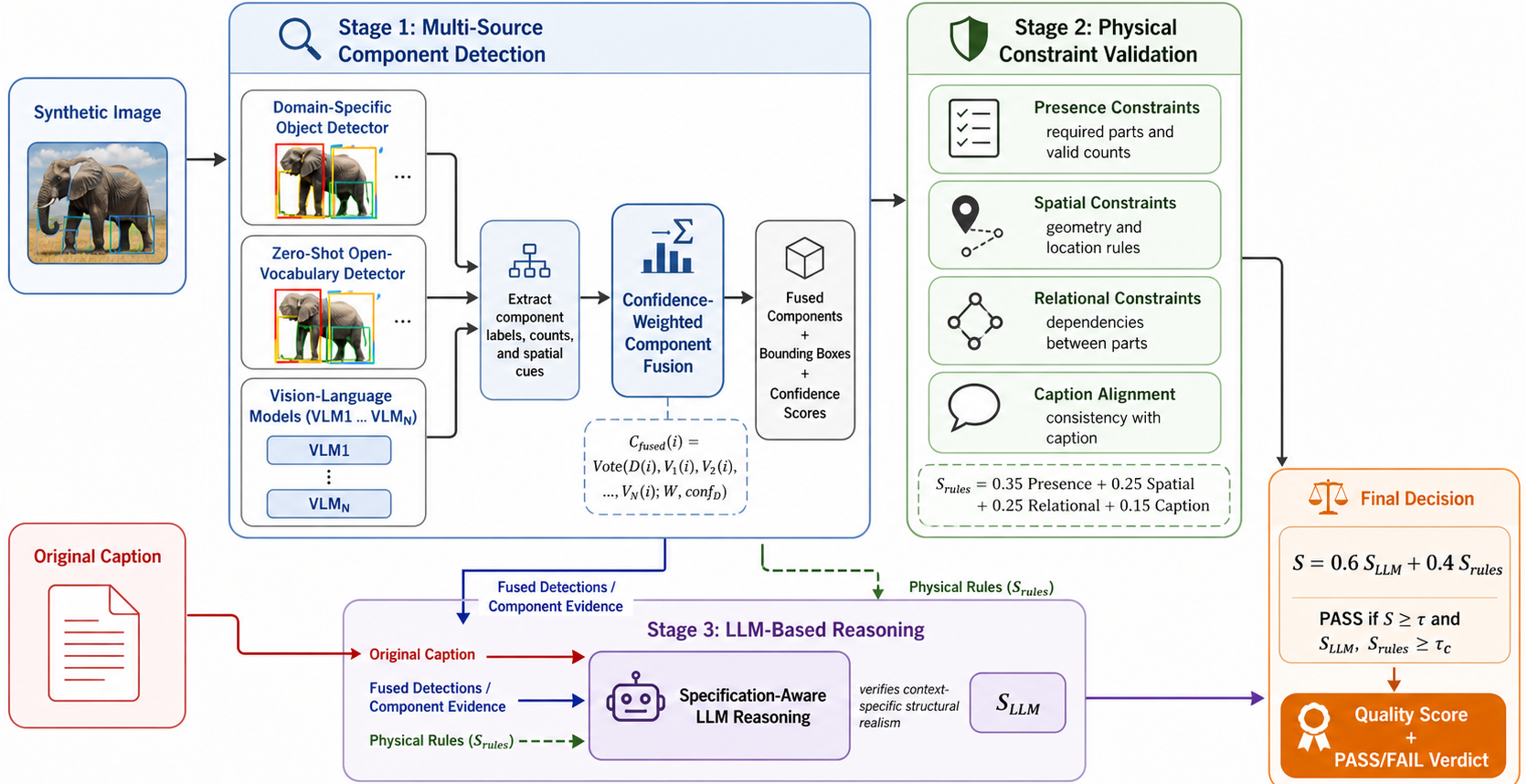}
\caption{\footnotesize Overview of the proposed hybrid evaluation pipeline combining Transformer model-based detection, vision-language models, and LLM reasoning with physical consistency rules.}
\label{fig:pipeline}
\end{figure*}
\vspace{-5pt}
\subsection{Problem Formulation}
\begin{algorithm}[t]
\scriptsize
\caption{\scriptsize Physical Realism Evaluation for Synthetic Images}
\label{alg:evaluation}
\begin{algorithmic}[1]
\REQUIRE Generated image $I$, caption $C$, contextual rules $\mathcal{R}$
\ENSURE 
  Realism score $S \in [0, 100]$, and \\
  verdict $V \in \{\text{PASS}, \text{FAIL}\}$

\STATE \textbf{Component/Object Detection:}
\STATE 
  $\{\mathbf{n}, \mathbf{B}, \rho_e\} \leftarrow \text{Object\_Fu\-sion}(I; \text{VIT}, \text{VLM}_n)$
  \COMMENT{Component counts, bounding boxes, confidence}

\STATE \textbf{LLM Knowledge Validation:}
\STATE $\mathcal{P} \leftarrow \text{FormatRules}(\mathbf{B}, \mathcal{R})$ 
\COMMENT{physics guided constraints with bounding box analysis}
\STATE $S_{\text{LLM}} \leftarrow \text{LLM}(C, \mathbf{n}, \mathbf{B}, \mathcal{P})$ 
\COMMENT{Specification + presence, spatial, relational validation}

\STATE \textbf{Deterministic Validation:}
\STATE $S_{\text{rules}} \leftarrow 0.35 \cdot \text{Presence}(\mathbf{n}) + 0.25 \cdot \text{Spatial}(\mathbf{B}, \mathcal{R})$
\STATE \hspace{3em} $+ 0.25 \cdot \text{Relational}(\mathbf{n}) + 0.15 \cdot \text{Caption}(\mathbf{n}, C)$

\STATE \textbf{Final Score:}
\STATE $S \leftarrow 0.6 \cdot S_{\text{rules}} + 0.4 \cdot S_{\text{LLM}}$
\STATE $V \leftarrow \text{PASS if } S \geq \tau \text{ and } S_{\text{LLM}}, S_{\text{rules}} \geq \tau_c$

\STATE \textbf{return} $S$, $V$
\end{algorithmic}
\end{algorithm}
\vspace{-5pt}

Given a synthetic image $I \in \mathbb{R}^{H \times W \times 3}$ and a corresponding text caption $C$, we aim to compute a physical realism score $S \in [0, 100]$ that quantifies whether the image depicts a physically plausible configuration. We denote the set of component classes as $\mathcal{C} = \{c_1, c_2, \ldots, c_M\}$, where $M$ is the total number of component types in the domain (e.g., $M=5$ for aircraft: head, tail, engine, wing, tail wing; $M=13$ for cars: wheels, bonnet, windshield, headlights, etc.; and $M=10$ for mammals: head, eye, ear, nose, horn, trunk, leg, tail, chest, and belly).

The challenge we address is that standard evaluation metrics, CLIPScore and SigLIP-2, measure only semantic alignment between image and text embeddings. These metrics assign high scores when objects mentioned in the caption appear in the image and are correctly classified. However, they cannot detect structural violations. For instance, an aircraft image with engines mounted upside-down on the wings receives high semantic alignment scores because both ``engines'' and ``wings'' are correctly identified in the image and mentioned in the caption, despite violating fundamental aerodynamic principles. Similarly, a car image with wheels positioned above the roofline receives high semantic scores because wheels are detected and the car is correctly identified, yet the alignment violates basic mechanical constraints where wheels must contact the ground to support the vehicle. In the animal domain, a generated image might depict a dog with wings or a cow with six legs; while the labels ``dog'' and ``legs'' are correctly identified, the resulting anatomical configuration is biologically impossible.

Our framework solves this by explicitly verifying three categories of physical constraints: \textbf{Presence Constraints}: Critical components must exist in realistic quantities (e.g., an aircraft requires exactly one head, one tail, and 2-4 engines; a car requires exactly one bonnet, windshield, and roof; and an animal requires exactly one head and one torso). Second \textbf{Spatial Constraints}: Components must occupy physically valid positions relative to each other (e.g., engines must be positioned near the wings for aircraft, not embedded in the fuselage; wheels must be at the bottom for cars, not above the roof; and eyes must be contained within the head for animals, not floating on the torso). At last \textbf{Relational Constraints}: Components must satisfy logical dependencies (e.g., engines cannot exist without wings to mount on for aircraft; wheels cannot exist without a vehicle body for cars; and eyes cannot exist without a corresponding head for animals). An image is physically questionable if it violates any of these constraints, regardless of how semantically aligned it is with the caption. Our framework detects such violations through a three-stage pipeline: (Stage 1) Multimodal component detection, (Stage 2) Rule based constraint validation, and (Stage 3) LLM based contextual reasoning.

\vspace{-10pt}
\subsection{Stage 1: Multi Source Component Detection and Fusion}
\vspace{-7pt}
Single detection models often fail to identify all components reliably, especially small objects like engines in aircraft images or side mirrors in car images. We employ multiple complementary detectors; a domain specific object detector and multiple general purpose VLMs, that together provide more robust component identification. The key insight is that different models excel at different types of observations: the object detector provides precise spatial localization, while VLMs extract semantic relationships and component counts from visual features.

\textbf{Custom Domain-Specific Detector (CDSD):}
For each critical domain (aircraft, cars, animal), we fine-tune a vision transformer-based object detection model to identify critical components. Vision transformer-based models offer strong performance on object detection with superior handling of spatial relationships and multi-scale objects. For \textbf{aircraft images}, we define five essential components: head (cockpit/nose section including the forward fuselage), engine (turbofan nacelles where jet fuel burns), swept wing (main lifting surfaces), tail (vertical stabilizer for directional control), and tail wing (horizontal stabilizer for pitch control). For \textbf{car images}, we define thirteen components: wheels, bonnet (front hood), windshield, headlights, taillights, doors, grille (front panel), trunk (rear compartment), roof, bumpers (front and rear), fenders (wheel housings), side windows, and rearview mirrors. The trained detector $D_\theta$ produces component detections
$D_\theta(I)=\{(b_i,c_i,\sigma_i)\}_{i=1}^{N}$, where
$b_i=[x_1,y_1,x_2,y_2]$ is a pixel-space bounding box,
$c_i\in\mathcal{C}$ is the predicted component class, and
$\sigma_i\in[0,1]$ is the confidence score. Detections are filtered using
$\sigma_i\geq0.5$, retaining high-confidence predictions and reducing false positives.

\textbf{Zero-Shot Open-Vocabulary Detector:}
For general domain datasets such as animals, we employ a zero-shot open-vocabulary detection model that does not require any domain-specific training. Unlike the custom detector, this model operates using free-form text prompts to identify components, allowing it to generalize across different object categories. The key idea is that the model leverages vision-language alignment to detect components based on their textual descriptions rather than fixed class labels, making it suitable for open-domain scenarios. Given an input image $I$, we first perform sub-domain classification by computing the alignment score between the image and a set of candidate sub-domain labels (e.g., mammal, bird). The predicted sub-domain $\hat{d}$ is selected as the label with the highest alignment score. If no sub-domain satisfies a minimum confidence threshold, we fall back to a generic prompt that includes all possible component categories to avoid missing detections. Based on the predicted sub-domain, we construct structured text prompts that describe relevant components (e.g., head, eye, leg, tail). These prompts are grouped into smaller sets to improve detection quality and are processed independently. The model then produces detections:
\begin{equation}
D_{\text{zero-shot}}(I) = {(b_i, c_i, \sigma_i)}_{i=1}^{N'}
\end{equation}
where $b_i$ is the bounding box, $c_i$ is the component class obtained through phrase matching, $\sigma_i$ is the confidence score, and $N'$ is the number of detections.

To improve robustness, we perform multi-scale inference and apply confidence-based filtering to remove low-quality detections. We further refine the results using non-maximum suppression and simple geometric constraints to eliminate duplicate or unrealistic component predictions.

\textbf{VLM Feature Extraction:} To complement the detector's spatial precision, we extract semantic descriptions and component relationships using multiple VLMs. Semantic understanding refers to the ability to recognize and interpret the meaning and relationships of visual objects in images. These models provide complementary semantic understanding without additional training. The vision-language models produce outputs that we parse to extract:
\begin{equation}
V_k(I) = \{(o_j^k, r_j^k)\}_{j=1}^{M_k}
\end{equation}
where $o_j^k$ is a semantic object description (e.g., ``four engines,'' ``swept-back wings'' for aircraft; ``four wheels,'' ``round headlights'' for cars; ``four legs,'' ``two wings'' for animals), $r_j^k$ is a relational statement (e.g., ``engines positioned beneath wing centerline'' for aircraft; ``wheels at bottom of vehicle body'' for cars; ``legs attached below torso centerline'' for animals), $k$ indexes the VLM and $M_k$ is the number of objects identified by the model. We parse these outputs using regular expression matching to extract structured information: component types, counts and spatial relationships. For instance, the phrase ``four engines below wings'' is parsed into: component\_type=``engine'', count=4, spatial\_relation=``below'', reference=``wings'' for aircraft; similarly, ``four wheels at bottom'' for cars is parsed as: component\_type=``wheel'', count=4, spatial\_relation=``at\_bottom'', reference=``vehicle\_body''; and ``four legs below torso'' for animals is parsed as: component\_type=``leg'', count=4, spatial\_relation=``below'', reference=``torso''.

\textbf{Confidence Weighted Fusion:} We fuse detections from all sources (vision transformer detector and multiple VLMs) using confidence weighted agreement. For each component class $c$, we aggregate detections through the following procedure:

\setlength{\abovedisplayskip}{3pt}
\setlength{\belowdisplayskip}{3pt}
\setlength{\abovedisplayshortskip}{2pt}
\setlength{\belowdisplayshortskip}{2pt}

\noindent\textbf{Step 1: Geometric Agreement.}
We compute intersection over union (IoU) between bounding boxes from different models,
$\mathrm{IoU}(b_i,b_j)=
\frac{|b_i\cap b_j|}{|b_i\cup b_j|}$.
Two detections are treated as the same object when $\mathrm{IoU}(b_i,b_j)>0.5$, where the threshold allows small localization differences while still requiring sufficient overlap.

\noindent\textbf{Step 2: Semantic Agreement.}
We match VLM-extracted component descriptions to detector classes using keyword matching. For example, ``four engines'' maps to the engine class with count $=4$ for aircraft; ``four wheels'' maps to the wheel class with count $=4$ for cars; and ``four legs'' maps to the leg class with count $=4$ for animals.

\noindent\textbf{Step 3: Confidence Weighting.}
Each detection $d_i$ from model $m$ receives a normalized confidence weight,
$w_i=\sigma_i^m/\sum_j\sigma_j^m$,
where $\sigma_i^m$ is the confidence from model $m$. The normalized weights sum to one within each component class.

\noindent\textbf{Step 4: Final Count Aggregation.}
For component class $c$, the final count is computed as a weighted median across detector and VLM estimates:
\[
N_c^{\mathrm{final}} =
\operatorname{wmed}\!\left(
\{N_c^{\mathrm{det}},N_c^{\mathrm{VLM1}},N_c^{\mathrm{VLM2}}\},
\{w_{\mathrm{det}},w_{\mathrm{VLM1}},w_{\mathrm{VLM2}}\}
\right).
\]
This fusion reduces sensitivity to outlier predictions. If model counts differ by more than one component, e.g., $2$ vs. $4$ engines, $4$ vs. $3$ wheels, or $4$ vs. $6$ legs, the image is flagged for manual inspection, while majority agreement is used to prevent cascading failures. The fused output is
\[
O_f=\{(c,N_c^{\mathrm{final}},B_c,\rho_c):c\in\mathcal{C}\},
\]
where $N_c^{\mathrm{final}}$ is the final count, $B_c=\{b_i\}_{i=1}^{N_c^{\mathrm{final}}}$ are fused boxes, and $\rho_c$ is the fusion confidence.
\vspace{-5pt}
\subsection{Stage 2: Rule Based Physical Constraint Validation}
While automated detection identifies components and their locations, validating whether these arrangements satisfy physical principles requires encoding domain knowledge as logical constraints. We organize these constraints into three categories; presence, spatial and relational, that together capture the essential physics of structurally valid aircraft, vehicle designs, and animal anatomy. This approach is interpretable: when an image fails validation, specific violated rules can be reported to the user.

\textbf{Presence Rules:} These rules verify that mandatory components exist in realistic quantities based on domain knowledge. \textit{For aircraft}: (P1) One head, (P2) one tail, (P3) 2--4 engines, (P4) 1--2 wings, (P5) engine width in range $[0.05 L_{\text{fus}}, 0.15 L_{\text{fus}}]$. \textit{For cars}: (P1) wheels in $\{2, 4, 6\}$, (P2) exactly one bonnet, windshield, roof, (P3) headlights/taillights in even pairs. \textit{For animals}: (P1) Exactly one head and torso, (P2) leg count matching subtype (4 for quadrupeds, 2 for bipeds), (P3) wings and beaks restricted to avian subtypes, (P4) zero foreign anatomical features. Presence score: $s_{\text{pres}} = \frac{\text{satisfied rules}}{\text{total rules}}$.

\textbf{Spatial Rules:} These rules enforce that component bounding boxes satisfy geometric relationships derived from physical feasibility. \textit{For aircraft}: (S1) Engines near wings ($d_y < 50$ px or within expanded wing region), (S2) head left of tail, (S3) tail at rear ($x > 0.7W$), (S4) wings intersect centerline. \textit{For cars}: (S1) Wheels at bottom ($y > 0.6H$), (S2) roof at top ($y < 0.4H$), (S3) windshield between bonnet and roof, (S4) lights positioned correctly. \textit{For animals}: (S1) Eyes and beak contained within head (containment $\geq 0.70$), (S2) legs attached below torso, (S3) legs do not overlap head ($IoU \leq 0.50$), (S4) head at anterior end. Spatial score: $s_{\text{spat}} = \frac{\text{satisfied rules}}{\text{total rules}}$.

\textbf{Relational Rules:} These rules encode logical dependencies: certain components can only exist if other components are present. \textit{For aircraft}: (R1) Engines $\Rightarrow$ wings, (R2) head $\Leftrightarrow$ tail, (R3) engine/fuselage ratio in $[0.05, 0.15]$. \textit{For cars}: (R1) Wheels $\Rightarrow$ body, (R2) headlights $\Leftrightarrow$ taillights, (R3) wheel/vehicle ratio in $[0.10, 0.25]$. \textit{For animals}: (R1) Legs $\Rightarrow$ torso, (R2) eyes $\Rightarrow$ head, (R3) torso as the largest component by area, (R4) head/torso width ratio in $[0.15, 0.50]$. Relational score: $s_{\text{rel}} = \frac{\text{satisfied rules}}{\text{total rules}}$.

\textbf{Caption Alignment:} Beyond physical validity, the generated image must also be consistent with the textual description and the subtype. We define this category as a type-specific consistency score:
\begin{equation}
s_{\text{type}} = \frac{\text{number of satisfied caption- and subtype-specific rules}}{\text{total caption- and subtype-specific rules}}.
\end{equation}
This score checks whether the components mentioned in the caption are present in the image and whether they are compatible with the predicted subtype. For example, if the caption states ``a car with four wheels,'' we verify that wheels are detected in the image, allowing limited tolerance for occlusion and viewpoint variation so that at least three wheels may still be accepted as visible. Similarly, for captions such as ``a DC-10 aircraft,'' we check whether the detected aircraft components are consistent with a tri-jet configuration. In animal domains, the same category also covers subtype-dependent rules, such as allowing wings and beaks for birds while requiring them to have zero count for other subtypes. Limited tolerance is allowed to account for partial visibility and perspective effects.

\textbf{Rule-Based Score Aggregation:} The four constraint categories are combined using a weighted summation,
$S_{\text{rules}} = 100 \cdot (w_p s_{\text{pres}} + w_s s_{\text{spat}} + w_r s_{\text{rel}} + w_c s_{\text{cap}})$,
where $w_p = 0.35$, $w_s = 0.25$, $w_r = 0.25$, and $w_c = 0.15$. The weights were determined through a preliminary study involving 70 images annotated by three domain experts, selected to maximize Pearson correlation with expert ratings ($\rho = 0.79$). The presence constraint receives the highest weight because component existence is foundational, while spatial and relational constraints capture geometric and logical feasibility. Caption alignment is weighted lowest, as valid configurations may omit explicit mention of all components.
\vspace{-10pt}
\subsection{Stage 3: LLM Based Contextual Reasoning}
While rules capture geometric and logical constraints, specification aware reasoning; matching the image to specific aircraft, vehicle, or animal types mentioned in the caption, requires semantic understanding beyond fixed predicates. An aircraft labeled ``DC-10'' must have exactly 3 engines (tri-jet configuration), while a ``Boeing 747'' requires 4 engines. A car labeled ``sedan'' should have four doors, while a ``coupe'' typically has two. Similarly, an animal labeled ``dog'' requires 4 legs and no wings, whereas a ``chicken'' requires 2 legs and wings. These type specific constraints cannot be efficiently encoded as rules because they depend on understanding natural language descriptions and domain knowledge that constantly evolves. We employ a LLM to perform this reasoning. The LLM does not directly rescore the raw image; instead, it reasons over detector outputs, VLM-extracted observations, bounding-box geometry, and domain rules.

\textbf{LLM Prompt Construction:} The LLM receives a comprehensive prompt structured across four key components: \textit{(1) Image State}: Original caption $C$, detected components $O_f$ with counts and bounding boxes, spatial observations from VLMs. \textit{(2) Domain Knowledge}: Aircraft type specifications (e.g., DC-10 tri-jet with 3 engines, Boeing 747 quad-jet with 4 engines), vehicle type specifications (e.g., sedan with 4 doors and 4 wheels), and animal type specifications (e.g., dog with 4 legs and no wings, chicken with 2 legs and wings). \textit{(3) Evaluation Task}: Three explicit questions: (i) Does component count match the type specification? (ii) Are spatial relationships correct for this type? (iii) Are there physical impossibilities? \textit{(4) Scoring Rubric}: 0--30 (critical violations: wrong type, impossible configuration), 40--60 (major issues: count mismatch, incorrect placement), 70--85 (minor issues: plausible but unusual), 86--100 (valid: matches specifications).

\noindent The LLM generates score $S_{\text{LLM}} \in $ and textual explanation identifying specific violations. For concrete examples: aircraft caption specifies ``DC-10'' (tri-jet), but detected configuration shows 4 engines → critical violation → score $\approx 25$; car caption specifies ``sedan,'' detected configuration has 4 doors and 4 wheels matching sedan specifications → valid configuration → score $\approx 92$; animal caption specifies ``dog,'' but detected configuration shows wings → critical violation → score $\approx 25$; animal caption specifies ``chicken,'' detected configuration has 2 legs and wings matching chicken specifications → valid configuration → score $\approx 92$.

\textbf{LLM Scoring Procedure:} We use a LLM with strong reasoning capabilities for specification aware validation. The LLM generates both a numerical score $S_{\text{LLM}} \in $ and a textual explanation $E_{\text{LLM}}$ that identifies specific violations. To ensure reliability in the face of LLM stochasticity, we run the evaluation three times per image:
\begin{equation}
(S_{\text{LLM}}^{(1)}, S_{\text{LLM}}^{(2)}, S_{\text{LLM}}^{(3)}) = \text{LLM}(\text{Prompt})
\end{equation}
If the variance of these three scores exceeds a threshold (10 points), we rerun until convergence or take the median of three runs. This procedure prevents single run outliers from affecting the final score. We use sampling temperature $\tau = 0.3$ for more deterministic outputs and limit generation to max\_tokens = 500 to keep explanations concise.

\textbf{Handling Ambiguous Cases:} When the caption is ambiguous or underspecified (e.g., just ``an aircraft'' without type, ``a car'' without specification, or ``an animal''), the LLM assesses whether the configuration is plausible for any realistic domain type. For aircraft, this might mean checking if the engine count matches common aircraft configurations (2 or 4 for commercial, 3 for tri-jets). For cars, this means verifying typical vehicle proportions and component arrangements. For animals, this means verifying expected anatomical structures like 4 legs for common mammals. This prevents over penalizing valid configurations that happen to deviate from under specified captions.
\vspace{-13pt}
\subsection{Hybrid Score Fusion and Decision Logic}
\vspace{-7pt}
\textbf{Final Score Computation:} Rule-based and LLM-based scores are combined using fixed weights,
$S_{\text{final}} = 0.6\, S_{\text{rules}} + 0.4\, S_{\text{LLM}}$,
where the weighting balances deterministic, interpretable geometric validation from rules with adaptive, knowledge-based reasoning from the LLM.

\textbf{Pass/Fail Decision with Dual Thresholds:} An image is assigned a PASS verdict only if
$S_{\text{final}} \ge \tau$ and $S_{\text{rules}} \ge \tau_c$ and $S_{\text{LLM}} \ge \tau_c$,
with $\tau = 60$ and $\tau_c = 40$. The dual-threshold criterion prevents compensation effects, where a high score in one component could mask a critical failure in the other. For example, an image may achieve $S_{\text{rules}} = 95$ due to correct geometric structure but only $S_{\text{LLM}} = 30$ because of an incorrect object specification, resulting in $S_{\text{final}} = 69.0$. Although the final score exceeds $\tau$, the image correctly fails due to $S_{\text{LLM}} < \tau_c$. Similar failure modes occur for vehicle images with valid spatial layouts but mismatched types, or animal images that satisfy generic quadrupedal rules but violate species-specific specifications, such as a dog depicted with wings or an elephant exhibiting anatomical merging between overlapping subjects.

\textbf{Threshold Justification:} Thresholds $\tau$ and $\tau_c$ were selected using grid search on a held-out validation set consisting of 70 images per domain. Each image was independently labeled by three domain experts as either \emph{physically correct} or \emph{physically incorrect}. Inter-annotator reliability was assessed using Fleiss' kappa, yielding $\kappa = 0.81$, indicating substantial agreement. A grid search was conducted over $\tau \in \{40, 50, 60, 70, 80\}$ and $\tau_c \in \{30, 40, 50, 60\}$. For each threshold pair, precision, recall, and F1-score were computed. The optimal thresholds, $\tau = 60$ and $\tau_c = 40$, were selected as they maximized the F1-score ($\text{F1} = 0.87$, $\text{Precision} = 0.89$, $\text{Recall} = 0.85$). These metrics indicate that at the chosen thresholds, the method correctly identifies $\sim 87\%$ of images as physically plausible or implausible, with 11\% false positive rate and 15\% false negative rate.

\textbf{Diagnostic Output:} For FAIL verdicts, we output violation diagnostics $V_{\text{diag}} = \{V_{\text{spec}}, V_{\text{spatial}}, V_{\text{rules}}\}$ identifying specification violations (e.g., wrong aircraft type, door count mismatch, species-specific limb count), geometric violations (e.g., component misalignment, malformed animal limbs) and logical violations (e.g., missing required components, anatomical merging). This enables systematic failure analysis.

Our three stage methodology; multi source detection, rule based validation and LLM reasoning, provides a comprehensive framework for physical realism assessment across structured domains. The framework is interpretable: each stage produces diagnostics explaining why an image passes or fails. The approach is domain adaptable: new domains require annotating $\sim 700$ images per domain for detector training and specifying physics rules specific to that domain, but no code changes are required. 

\begin{table*}[t]
\centering
\scriptsize
\setlength{\tabcolsep}{2.5pt}
\renewcommand{\arraystretch}{0.95}
\caption{\footnotesize Comprehensive metric statistics across 70 synthetic images per category. All metrics are reported on a common 0--100 scale; VQAScore, SigLIP-2, and FLEUR outputs (natively in $[0,1]$) are multiplied by 100 for comparability with PCMDE and CLIPScore.}
\label{tab:merged_metric_statistics_sidebyside}

\begin{minipage}[t]{0.49\textwidth}
\centering
\caption*{\footnotesize (a) Central tendency and score limits}
\vspace{-4pt}
\begin{tabular}{llcccc}
\toprule
\textbf{Cat.} & \textbf{Metric} & \textbf{Mean} & \textbf{Std} & \textbf{Min} & \textbf{Max} \\
\midrule
\multirow{5}{*}{Aircraft}
& \textcolor{blue}{\textbf{PCMDE}} & \textbf{74.8} & \textbf{12.67} & 50.0 & 92.5 \\
& CLIP     & 29.7 & 2.08 & 25.68 & 34.7 \\
& VQA      & 81.0 & 7.7 & 53.0 & 92.0 \\
& SigLip-2 & 99.9 & 0.2 & 99.0 & 100.0 \\
& FLEUR    & 76.8 & 4.6 & 63.7 & 84.5 \\
\midrule
\multirow{5}{*}{Car}
& \textcolor{blue}{\textbf{PCMDE}} & \textbf{79.0} & \textbf{9.6} & 46.0 & 91.6 \\
& CLIP     & 28.36 & 6.34 & 2.54 & 37.6 \\
& VQA      & 72.5 & 25.2 & 6.0 & 93.0 \\
& SigLip-2 & 99.0 & 1.7 & 89.0 & 100.0 \\
& FLEUR    & 73.8 & 14.1 & 8.7 & 83.9 \\
\midrule
\multirow{5}{*}{Animal}
& \textcolor{blue}{\textbf{PCMDE}} & \textbf{81.6} & \textbf{8.7} & 56.0 & 98.0 \\
& CLIP     & 32.8 & 2.5 & 26.7 & 40.3 \\
& VQA      & 77.1 & 7.8 & 48.3 & 89.8 \\
& SigLip-2 & 99.9 & 0.4 & 96.3 & 100.0 \\
& FLEUR    & 77.0 & 5.0 & 54.6 & 85.6 \\
\bottomrule
\end{tabular}
\end{minipage}
\hfill
\begin{minipage}[t]{0.49\textwidth}
\centering
\caption*{\footnotesize (b) Discriminative spread and variability}
\vspace{-4pt}
\begin{tabular}{llccc}
\toprule
\textbf{Cat.} & \textbf{Metric} & \textbf{Range} & \textbf{CV} & \textbf{Finding} \\
\midrule
\multirow{5}{*}{Aircraft}
& \textcolor{blue}{\textbf{PCMDE}} & \textbf{42.5} & \textbf{16.9} & Stable \\
& CLIP     & 9.04 & 7.0 & Low spread \\
& VQA      & 39.0 & 9.5 & Moderate \\
& SigLip-2 & 1.0 & 0.2 & Saturated \\
& FLEUR    & 20.8 & 6.0 & Limited \\
\midrule
\multirow{5}{*}{Car}
& \textcolor{blue}{\textbf{PCMDE}} & \textbf{45.6} & \textbf{12.15} & Stable \\
& CLIP     & 35.06 & 22.4 & Volatile \\
& VQA      & 87.0 & 34.8 & Volatile \\
& SigLip-2 & 11.0 & 1.7 & Saturated \\
& FLEUR    & 75.2 & 19.1 & Unstable \\
\midrule
\multirow{5}{*}{Animal}
& \textcolor{blue}{\textbf{PCMDE}} & \textbf{42.0} & \textbf{10.7} & Stable \\
& CLIP     & 13.6 & 7.6 & Low spread \\
& VQA      & 41.5 & 10.1 & Moderate \\
& SigLip-2 & 3.7 & 0.4 & Saturated \\
& FLEUR    & 31.0 & 6.5 & Limited \\
\bottomrule
\end{tabular}
\end{minipage}

\vspace{-1pt}
\caption*{\footnotesize \textit{PCMDE maintains strong discriminative spread with stable CV across all categories. SigLip-2 shows saturation, while CLIP, VQA, and FLEUR often show either restricted spread or unstable variability.}}
\vspace{-15pt} 
\end{table*}

\vspace{-15pt}
\section{Experimentation and Results}
\vspace{-10pt}

\textbf{Experimental Setup} We chose to work with three domains: a car image dataset~\cite{car_models_3887}, an aircraft image dataset~\cite{fgvc_aircraft_dataset}, and animal images. We identified eight specific subtypes within the animal domain: dog, cat, horse, sheep, cow, squirrel, elephant, and chicken. For the aircraft and car domains, we meticulously selected 700 images to ensure computational feasibility and statistical diversity, providing an adequate sample size for feature extraction and further inference. We manually annotated the car and aircraft images on Roboflow~\cite{roboflow_platform} and pretrained YOLOv12~\cite{yolo12} from the ground up. In the animal domain, we employed the Zero-Shot Open-Vocabulary Detector utilizing Grounding DINO~\cite{liu2024groundingdinomarryingdino} in conjunction with SAM~\cite{kirillov2023segment}, which capitalizes on established vision-language alignment and eliminates the necessity for domain-specific pretraining. These classes were chosen for each dataset so no physical aspects for any of the datasets get missed, and proper contextual rules that have been implemented in the framework can guide the evaluation process swiftly. All aggregation weights and decision thresholds were selected on a held-out validation split that was disjoint from the reported test images. The reported 70 images per domain were used only for final metric comparison, ablation analysis, and case-study diagnostics. This separation avoids tuning PCMDE on the same samples used to report final performance. \textbf{Comparison Methods:} 
 1) CLIPcore measures how similar an image and text are using CLIP~\cite{radford2021learningtransferablevisualmodels} embeddings, 2) VQAScore uses visual question answering to check how well captions match images, 3) SigLIP-2 is a better version of CLIP, made for vision-language tasks, and, 4) FLEUR~\cite{lee2024fleurexplainablereferencefreeevaluation} is an explainable reference-free metric that scores captions against images using a large multimodal model and gives a smooth continuous score. 

\textbf{Evaluation Metrics.} Three statistical metrics used to compare the performance of our framework PCMDE with other metrics. The used metrics are (1) \textit{Standard Deviation}, (2) \textit{Range} ($Max - Min$), which assesses the ability of a metric to differentiate between favorable and unfavorable samples, and (3) \textit{Coefficient of Variation} ($CV = Std/Mean \times 100\%$), which standardizes variability across disparate scales, allowing for an independent comparison between metrics with varying ranges ($CLIPScore \in [0,100]$ vs. $SigLIP\text{-}2 \in [0,1]$). We use CV as a descriptive measure of relative score variability rather than as a formal statistical significance test. Moderate CV values are interpreted together with range, case-study diagnostics, and agreement with human annotations. A notably low CV score ($CV \leq 5\%$) typically indicates score saturation, while a significantly high CV score ($CV \geq 30\%$) suggests inadequate scaling rather than meaningful sensitivity. The main purpose of PCMDE is not to maximize score variance, but to detect physically implausible configurations that remain semantically aligned with the caption. Therefore, we interpret range and CV only as descriptive evidence, and rely on human agreement (Appendix \ref{sec:human_agreement} ) and case-study diagnostics(Appendix \ref{lab:case}) as the primary validation signals.

\textbf{Comparison with State-of-the-art.}Table~\ref{tab:merged_metric_statistics_sidebyside} shows that existing semantic-alignment metrics provide limited structural discrimination. SigLIP-2 is strongly saturated across all domains, with near-maximal scores and very low variability (aircraft: CV = 0.2\%, range = 1.0; cars: CV = 1.5\%, range = 11.0; animals: CV = 0.4\%, range = 3.7). CLIPScore shows restricted spread for aircraft and animals, while its larger car-domain variability is mainly caused by outliers rather than stable quality separation. VQAScore is effective for coarse caption verification but remains insensitive to subtle physical violations such as invalid component counts or misplaced parts. FLEUR shows a similar limitation: it provides caption-level explanations but does not ground them in component-level physical constraints.
In contrast, PCMDE maintains consistent structural discrimination across all three domains, with CV values between 10.7\% and 16.9\% and absolute score ranges between 42.0 and 45.6. We interpret these statistics as descriptive evidence rather than the sole validation signal: the main goal of PCMDE is to detect physically implausible configurations that remain semantically aligned with the caption. To verify this, \ref{sec:human_agreement} reports that PCMDE matches majority-vote human annotations in 200/210 images, yielding 95.2\% agreement across aircraft, car, and animal domains.
The ablation results in \ref{sec:ablation}  further show that the full PCMDE pipeline is more reliable than individual modules. Single-detector variants substantially reduce pass-rate stability, while LLM-only and VLM-only variants are more likely to overestimate structurally flawed images in the case studies. These results support the hybrid design: deterministic physical rules provide stable component-level grounding, while LLM reasoning handles subtype-specific constraints that are difficult to encode exhaustively. \textbf{Positioning Relative to Existing Metrics.} We note that PCMDE is not meant to replace embedding-based metrics such as CLIPScore, SigLIP-2, or VQAScore. These metrics are essential for assessing the semantic alignment between captions and images, which serves as a distinct and complementary goal to structural validation. Rather, PCMDE addresses a particular deficit in current evaluation tools: physical and structural plausibility within domain-specific scenarios, and is most useful when used alongside, rather than in place of, semantic alignment metrics.  The discriminative gaps reported above should be interpreted as evidence of this complementary role and not as a claim of overall superiority.  In case-study ablations (Appendix \ref{tab:ablation_and_cross_sidebyside}), VLM-only and LLM-only judges often overestimated flawed images, such as the car with impossible wheel placement or the animal examples with anatomical defects. The full pipeline corrected these failures by anchoring visual reasoning in deterministic component-level constraints.

\vspace{-15pt} 
\section{Limitations \& Future Works}\vspace{-10pt} 
While PCMDE demonstrates discriminative power over embedding-based metrics, several limitations remain that provide opportunities for future research. \textbf{Computational Overhead:} The current framework relies on a multi-stage pipeline involving multiple heavy models, including vision transformers (CDSD), large multimodal models (VLMs), and LLMs for reasoning. This results in significantly higher latency and memory requirements compared to single-pass metrics like CLIPScore or SigLIP-2. Future work could explore distilling this multi-source knowledge into a more efficient, unified evaluation encoder to reduce inference time. \textbf{ Rule Engineering and Validation:}  We currently utilize structured template prompts to guide LLMs in drafting domain-specific rules (e.g., defining valid relative positions for new object types). However, these LLM-generated rules may need manual human validation to ensure accuracy and scientific correctness. Future research will focus on developing automated verification protocols to verify rules. \textbf{Static 2D Limitations:} PCMDE evaluates 2D projections of 3D objects. Errors caused by perspective distortion or extreme occlusion can sometimes trigger false failure diagnostics in the rule-based stage. Integrating 3D-aware priors or evaluating structural integrity across multiple synthetic viewpoints of the same generated object could improve robustness in these cases.
\textbf{Broader impact and responsible use} PCMDE can support quality control for synthetic images in domains where structural plausibility matters, including vehicle, aircraft, animal, and scientific-image generation. However, it should not be used as a standalone safety or correctness certificate. The score depends on detector reliability, visible 2D evidence, and the coverage of domain rules. False failures may occur under occlusion or unusual viewpoints, and false passes may occur when violations fall outside the encoded rule set. Human review remains necessary for high-stakes use.

\vspace{-15pt}
\section{Conclusion}\vspace{-10pt}
\label{sec:conclusion} We present PCMDE, evaluation method that checks how realistic synthetic images are by combining object detection, spatial and contextual rules, and simple reasoning with a LLM. Tested on synthetic aircraft, car and animal datasets, PCMDE can tell the difference between realistic and unrealistic images, without the score limits seen in regular vision-language metrics.


\begin{ack}
\end{ack}

\bibliographystyle{unsrtnat}
\bibliography{bibliography}

\newpage
\appendix

\section{Appendix}
\label{sec:appendix_physical_rules}
\small
\subsection{Implementation Details}
Using YOLOv12 for both the aircraft and car datasets, we resized the images to $640 \times 640$, set a batch size of 16, and trained for 150 epochs, allowing for early stopping with a patience of 50 epochs. We also applied augmentation on the images to ensure robustness against varying illumination (the HSV color space was applied) with rotation ($\pm 10^\circ$), translation (10\%), and scaling (50\%). For testing, we created a test set with 70 images for each of the aircraft, car, and animal datasets. We created 70 synthetic test images per domain using the HiDream-I1~\cite{hidreami1technicalreport} model for the test set. We did inference on all test data. Then we used Deepseek-vl2~\cite{wu2024deepseekvl2mixtureofexpertsvisionlanguagemodels} and Pixtral-12b~\cite{agrawal2024pixtral12b} to analyze the same data in the same way as the primary detectors. While performing the evaluation with the pipeline, we used Deepseek-r1 70b~\cite{deepseekai2025deepseekr1incentivizingreasoningcapability} from ollama to perform the LLM reasoning of our pipeline. More implementation details are provided in Appendix. We used two disjoint generated splits per domain: a validation split of 70 images per domain for threshold and weight selection, and a test split of 70 images per domain for the final results reported in Tables~1--10. No test image was used to tune thresholds, scoring weights, prompts, or rules. Human agreement in Table~9 is computed on the final test split. All the experiments were conducted on a high-performance compute node equipped with 16 NVIDIA A100-SXM4 GPUs, each with 80GB of VRAM interconnected via NVLink. 

\subsubsection{Reproducible Code link:} https://anonymous.4open.science/r/PCMDE-70F7/README.md.  More development environment
informations are provided there.
\subsubsection{Assets and licenses}. The released artifact consists of evaluation code, rule files, prompt templates, detector configuration files, and scripts for reproducing the reported scores. Existing datasets, models, and tools are credited in the references, and their licenses and access terms are documented in the anonymous repository. Redistribution of generated or annotated evaluation assets is provided only where permitted by the original source terms.
\subsection{Ablation Study}\vspace{-5pt}
\label{sec:ablation}
To verify the contribution of each module in PCMDE, we conducted an ablation study on the aircraft, car, and animal datasets using 70 images per domain. We evaluated five variants: the full pipeline, a rules-only variant, an LLM-only variant, a VLM-only detection variant, and a domain-specific detection variant (CDSD for aircraft/cars and Zero-Shot for animals). For the rules-only and LLM-only settings, we changed only the scoring weights in the pipeline. For the detection ablations, we re-ran the evaluation using only a single detection source.

\begin{table}[H]
\centering
\scriptsize
\setlength{\tabcolsep}{1.0pt}
\renewcommand{\arraystretch}{0.92}
\caption{Ablation results and cross-dataset agreement with the full pipeline.}
\label{tab:ablation_and_cross_sidebyside}

\begin{minipage}[t]{0.56\textwidth}
\centering
\caption*{\footnotesize (a) Ablation results}

\begin{tabular}{llcccc}
\toprule
\textbf{Data} & \textbf{Variant} & \textbf{Mean} & \textbf{Std} & \textbf{CV} & \textbf{Pass} \\
\midrule

\multirow{5}{*}{Aircraft}
& \textbf{Full Pipe.} & 74.8 & 12.67 & 16.9 & 80.0\% \\
& Rules Only & 79.4 & 25.2 & 31.8 & 80.0\% \\
& LLM Only & 68.8 & 12.4 & 18.1 & 70.0\% \\
& VLM-only Det. & 74.4 & 13.3 & 17.9 & 80.0\% \\
& CDSD-only Det. & 62.8 & 16.4 & 26.1 & 45.7\% \\

\midrule

\multirow{5}{*}{Car}
& \textbf{Full Pipe.}& 79.0 & 9.6   & 12.15 & 62.9\% \\
& Rules Only & 70.4 & 21.3 & 30.3 & 60.0\% \\
& LLM Only & 65.7 & 17.6 & 26.7 & 64.3\% \\
& VLM-only Det. & 69.9 & 15.0 & 21.4 & 78.6\% \\
& CDSD-only Det. & 57.1 & 19.0 & 33.3 & 41.4\% \\

\midrule

\multirow{5}{*}{Animal}
& \textbf{Full Pipe.} & 81.6 & 8.7   & 10.7 & 75.7\%  \\
& Rules Only & 84.2 & 16.5 & 19.6 & 78.6\% \\
& LLM Only & 75.3 & 10.2 & 13.5 & 68.6\% \\
& VLM-only Det. & 79.1 & 11.4 & 14.4 & 72.9\% \\
& Zero-Shot Det. & 68.4 & 14.8 & 21.6 & 51.4\% \\

\bottomrule
\end{tabular}
\end{minipage}
\hspace{-0.03\textwidth}
\begin{minipage}[t]{0.43\textwidth}
\centering
\caption*{\footnotesize (b) Agreement with full pipeline}

\setlength{\tabcolsep}{1.3pt}
\begin{tabular}{lcccccc}
\toprule
\textbf{Variant} & \textbf{Air $r$} & \textbf{Air \%} & \textbf{Car $r$} & \textbf{Car \%} & \textbf{Ani $r$} & \textbf{Ani \%} \\
\midrule
Rules Only & 0.888 & 100.0 & 0.954 & 97.1 & 0.912 & 98.6 \\
LLM Only & 0.362 & 90.0 & 0.699 & 81.4 & 0.448 & 84.3 \\
VLM-only Det. & 0.854 & 94.3 & 0.021 & 58.6 & 0.781 & 87.1 \\
Domain-only Det. & 0.268 & 65.7 & 0.661 & 75.7 & 0.552 & 71.4 \\
\bottomrule
\end{tabular}
\end{minipage}

\vspace{-2pt}
\caption*{\footnotesize \textit{Panel (a) reports ablation statistics using mean, standard deviation, coefficient of variation (CV), and pass rate. Panel (b) reports Pearson correlation $r$ and verdict agreement percentage with the full pipeline.}}
\vspace{-8pt}
\end{table}

Table~\ref{tab:ablation_and_cross_sidebyside} shows that the full PCMDE pipeline provides the most balanced performance across aircraft, car, and animal domains. Single-detector variants substantially reduce reliability: CDSD-only detection lowers pass rates for aircraft and cars, while Zero-Shot-only detection drops the animal pass rate to 51.4\%. The full pipeline remains especially stable on animals, achieving a mean score of 81.6 with the lowest CV of 10.7\%. Although VLM-only detection performs well on simpler aircraft images, it is less reliable for structurally complex car components. Cross-dataset agreement further shows that the rules-only variant aligns most closely with the full system, with $r>0.88$ and verdict agreement above 97\%, confirming that deterministic geometric constraints are central to PCMDE's reliability. Although rules-only agrees strongly with the full pipeline, the LLM stage is important for subtype-specific cases that require external knowledge, such as distinguishing a DC-10 tri-jet from a Boeing 747 quad-jet or recognizing species-specific anatomical constraints that are not fully captured by generic geometric rules.


\subsection{Case Study}
\label{lab:case}
\textbf{Diagnostic Analysis.} Tables~\ref{tab:aircraft_diagnostic}, \ref{tab:car_diagnostic}, and \ref{tab:animal_diagnostic} compare metric outputs on representative images exhibiting physical inconsistencies or structural uniformity across all three domains. PCMDE clearly differentiates between structurally valid and implausible images. Unrealistic aircraft images (DC-10 with an engine above the wing, score 50.0; Embraer with an engine under the fuselage, score 57.5), implausible car images (side view car without doors and anatomically impossible wheel placement, score 48.0; side view car with all wheels and bumpers missing and the body hanging above the ground, score 46.0), and structurally inconsistent animal images (elephant with its trunk merging into its offspring, score 42.2; cow with a missing leg, score 51.8) all receive substantially lower scores (42.2-57.5) than their structurally consistent counterparts (80.6-91.6). This method naturally yields a 25-49 point separation across all three domains, with no post-hoc tuning of thresholds. In contrast, SigLip-2 shows near-complete saturation (0.99–1.0) for all twelve samples, which makes it insensitive to quality differences across all domains. Both VQAScore and CLIPScore distributions overlap heavily. VQAScore ranges from 0.74 to 0.86 in the plausible group and from 0.28 to 0.85 in the implausible group, showing no distinct separation pattern. In contrast, CLIP ranges from 24.1 to 31.6 in the implausible group and from 15.8 to 36.2 in the plausible group, notably assigning a score of 15.76 to a structurally complete``BMW X5’’. PCMDE is the sole methodology that integrates both discriminability and explainability, providing explicit elucidations that highlight specific concerns such as ``engine positioned above wing’’ for aircraft, ``missing wheels for side view’’ for automobiles, or ``anatomical merging between subjects’’ for animals. The lucid explanation ties the scores to real world rules, and therefore makes effective quality control possible in many areas where simple similarity scores cannot.

The per-variant ablation analysis for each domain is provided below.
\begin{figure}[H]
    \centering
    \begin{subfigure}[t]{0.24\textwidth}
        \centering
        \includegraphics[width=\linewidth]{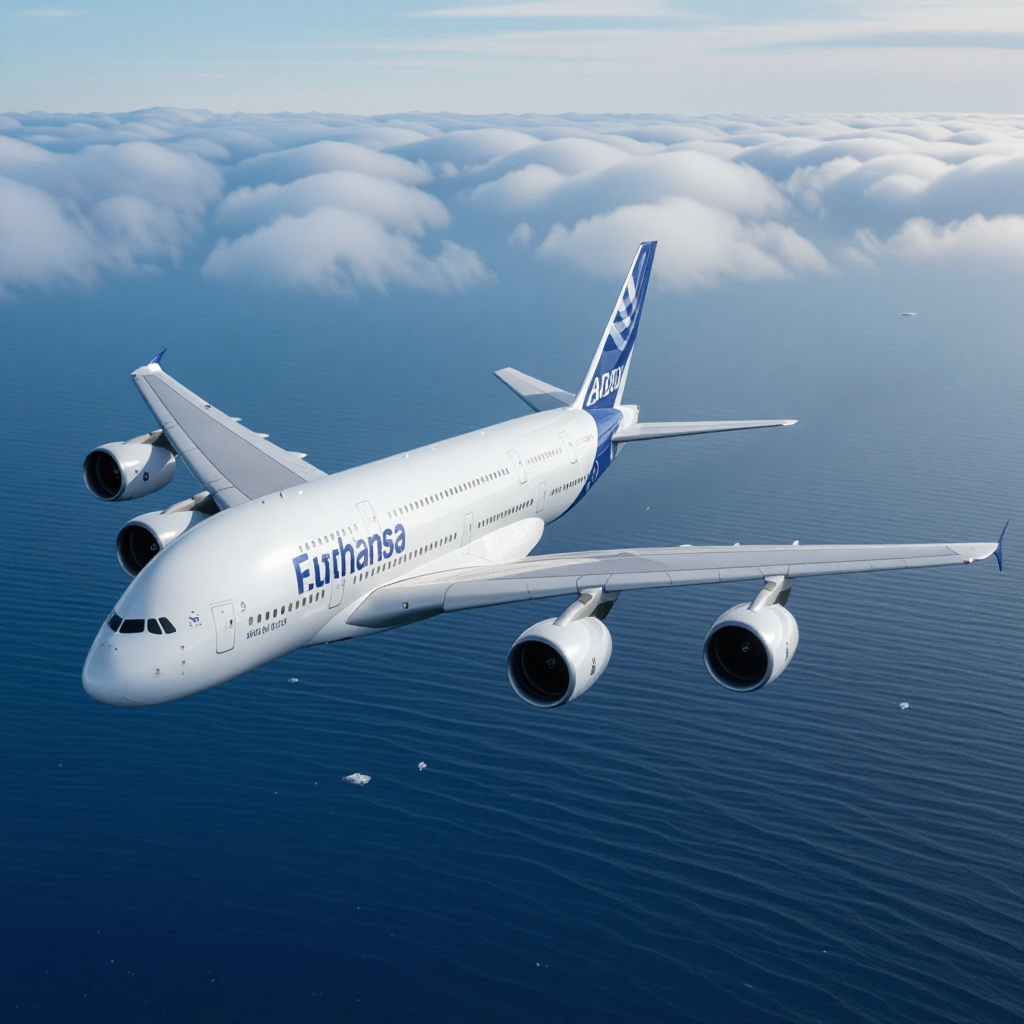}
        \caption{Structurally consistent: all components present.}
        \label{fig:aircraft_good1}
    \end{subfigure}
    \hfill
    \begin{subfigure}[t]{0.24\textwidth}
        \centering
        \includegraphics[width=\linewidth]{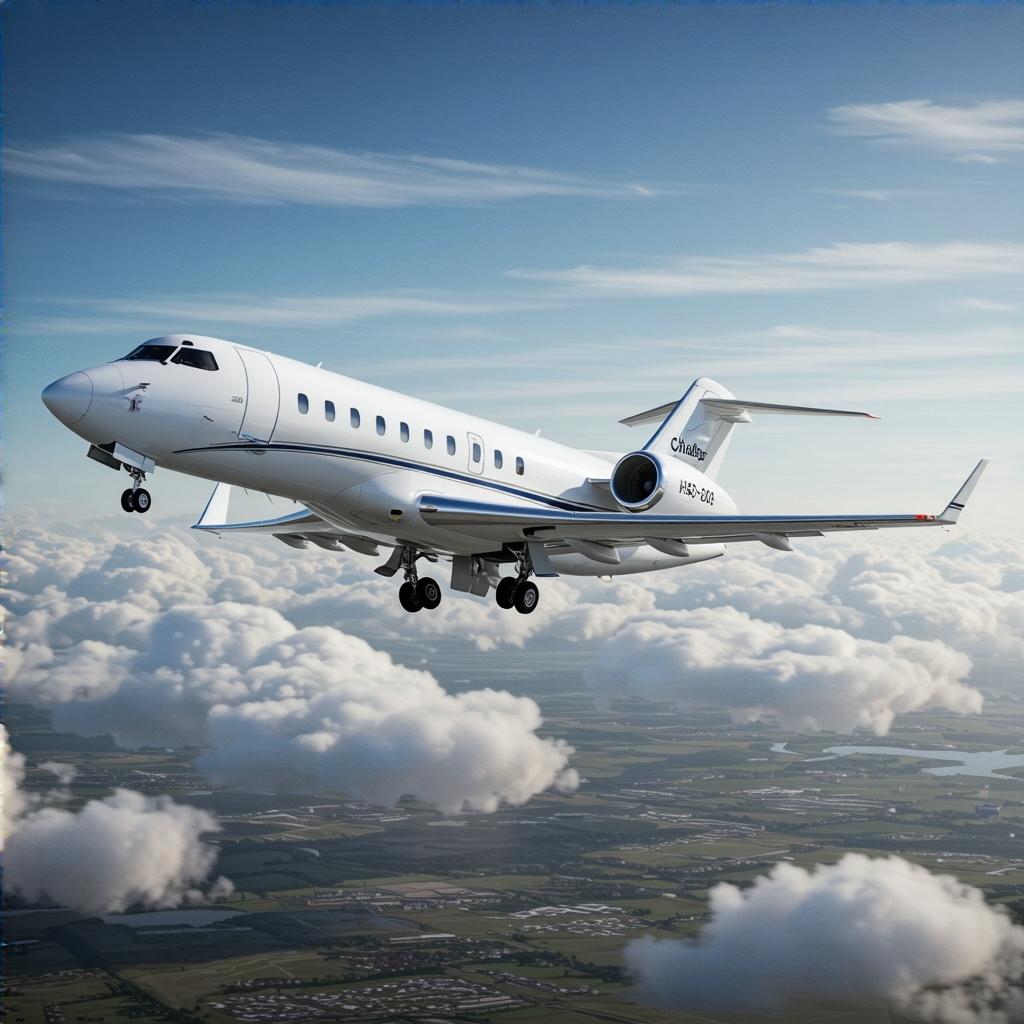}
        \caption{Physically plausible with no spatial violation.}
        \label{fig:aircraft_good2}
    \end{subfigure}
    \hfill
    \begin{subfigure}[t]{0.24\textwidth}
        \centering
        \includegraphics[width=\linewidth]{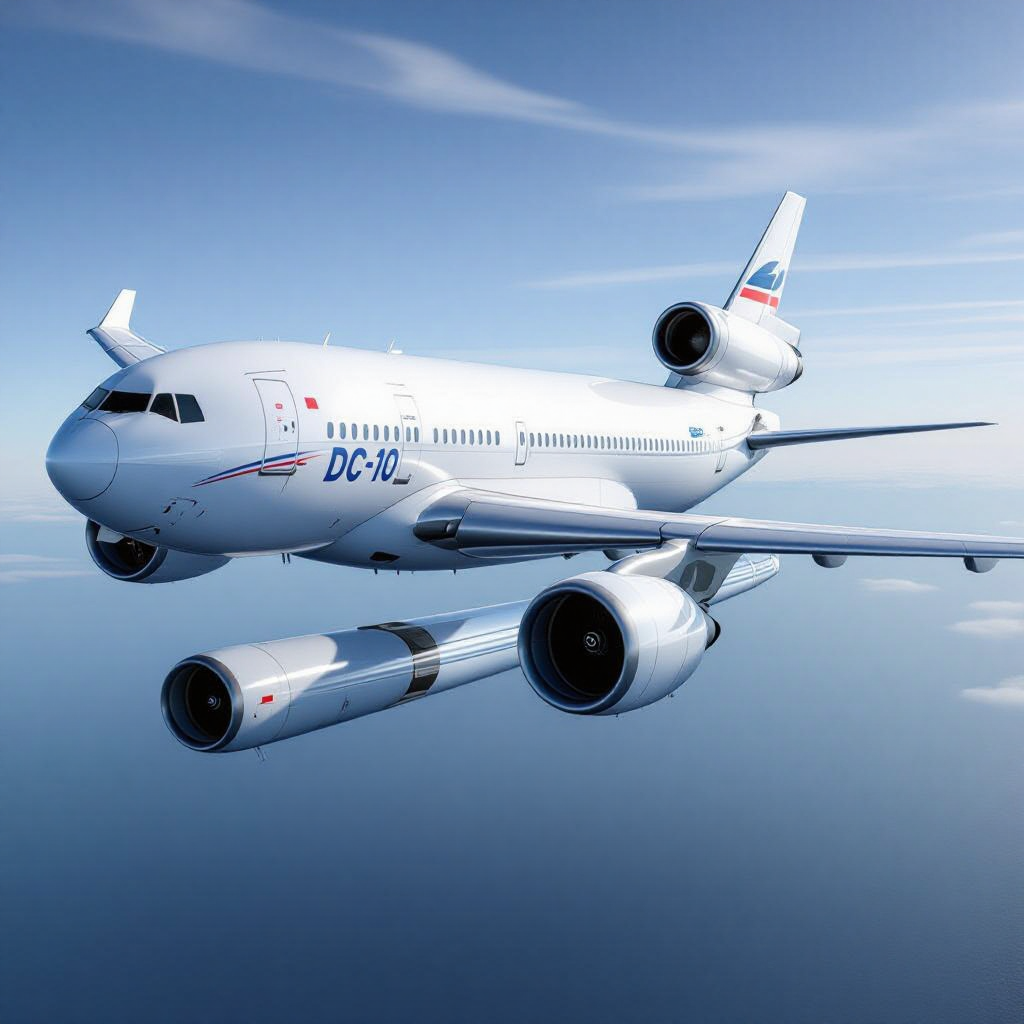}
        \caption{Implausible: 4 engines; one extra under wing.}
        \label{fig:aircraft_bad1}
    \end{subfigure}
    \hfill
    \begin{subfigure}[t]{0.24\textwidth}
        \centering
        \includegraphics[width=\linewidth]{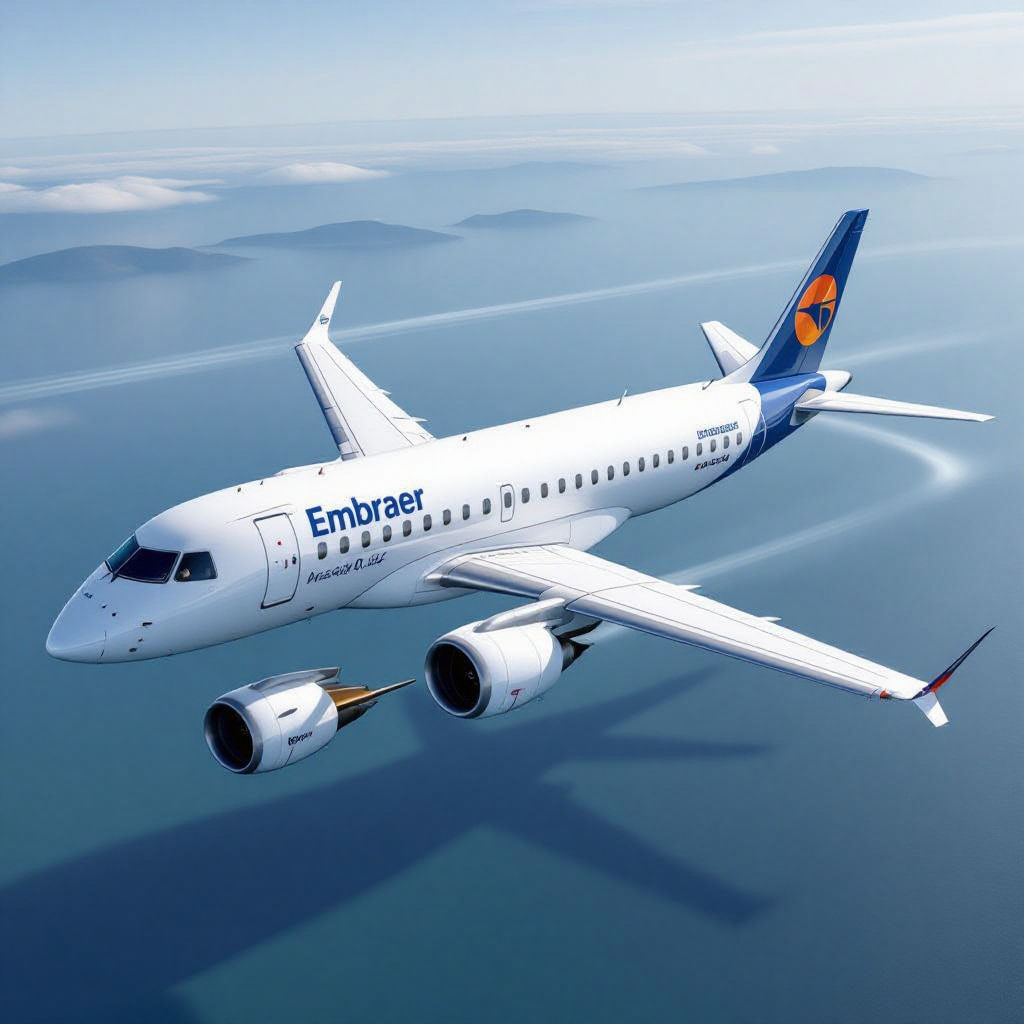}
        \caption{Extra wing and engine under fuselage.}
        \label{fig:aircraft_bad2}
    \end{subfigure}

    \caption{Illustrations of structurally sound and physically impossible aircraft.}
    \label{fig:sample_aircraft}
\end{figure}

\begin{table}[H]
    \centering
    \scriptsize
    \caption{\textbf{Metric scores on aircraft images with varying physical plausibility}}
    \label{tab:aircraft_diagnostic}
    \begin{tabularx}{\textwidth}{lccccX}
        \toprule
        \textbf{Image} & \textbf{PCMDE} & \textbf{CLIP} & \textbf{VQA} & \textbf{SigLip-2} & \textbf{Observed Physical Characteristics} \\
        \midrule
        \textbf{(a)} & 91.4 & 31.27 & 0.77 & 1.00 & Structurally consistent: All components present, engines correctly positioned. \\
        \textbf{(b)} & 90.0 & 32.50 & 0.84 & 1.00 & Physically plausible generic aircraft: No spatial violations or component miss. \\
        \textbf{(c)} & 50.0 & 30.35 & 0.81 & 1.00 & DC-10 with four engines (expected three); one engine positioned under wing. \\
        \textbf{(d)} & 57.5 & 30.15 & 0.86 & 0.99 & Embraer E-170 with an engine misplaced under fuselage. \\
        \bottomrule
    \end{tabularx}
\end{table}

\begin{table}[H]
\centering
\caption{PCMDE vs. Model-as-Judge baselines for aircraft images}
\label{tab:aircraft_ablation} 
\scriptsize
\begin{tabular}{lccccc}
\toprule
\textbf{Image} & \textbf{PCMDE} & \textbf{LLM-only} & \textbf{VLM-only} & \textbf{YOLO-only} & \textbf{Rules-only} \\
\midrule
(a) & 91.4 & 90.0 & 91.4 & 40.0 & 95.5 \\
(b) & 90.0 & 70.0 & 85.5 & 50.0 & 100.0 \\
(c) & 50.0 & 20.0 & 40.0 & 50.0 & 30.0 \\
(d) & 57.5 & 85.0 & 50.0 & 50.0 & 30.0 \\
\bottomrule
\end{tabular}
\end{table}
\textbf{Interpretation of Results:}  As shown in Table~\ref{tab:aircraft_ablation},, images (a) and (b) are correctly treated as structurally valid by the integrated PCMDE pipeline, receiving scores of 91.4 and 90.0, respectively. These scores are consistent with the Rules-only scores of 95.5 and 100.0, indicating that the deterministic physical constraints strongly support the plausibility of both aircraft. VLM-only also assigns high scores to these two plausible cases (91.4 and 85.5), while LLM-only is strong on image (a) but more conservative on image (b), assigning 90.0 and 70.0, respectively. In contrast, YOLO-only assigns low scores to the same plausible images (40.0 and 50.0), showing that detector-only evidence is insufficient without rule-based and semantic reasoning. Images (c) and (d) contain physical inconsistencies and receive lower PCMDE scores of 50.0 and 57.5. For image (c), the DC-10 configuration includes four engines instead of the expected three, and the Rules-only and VLM-only variants assign low scores of 30.0 and 40.0. The LLM-only score is even lower at 20.0, while YOLO-only gives 50.0 because it detects components but does not fully reason over aircraft-type constraints. Image (d) exposes a complementary failure mode: LLM-only assigns a high score of 85.0 despite the engine-placement violation, whereas Rules-only correctly penalizes the image at 30.0. The full PCMDE score of 57.5 reflects this conflict by combining semantic reasoning with deterministic spatial validation, allowing the pipeline to flag the image as structurally questionable rather than accepting the LLM-only judgment.
\begin{figure}[H]
    \centering

    \begin{subfigure}[t]{0.24\textwidth}
        \centering
        \includegraphics[width=\linewidth]{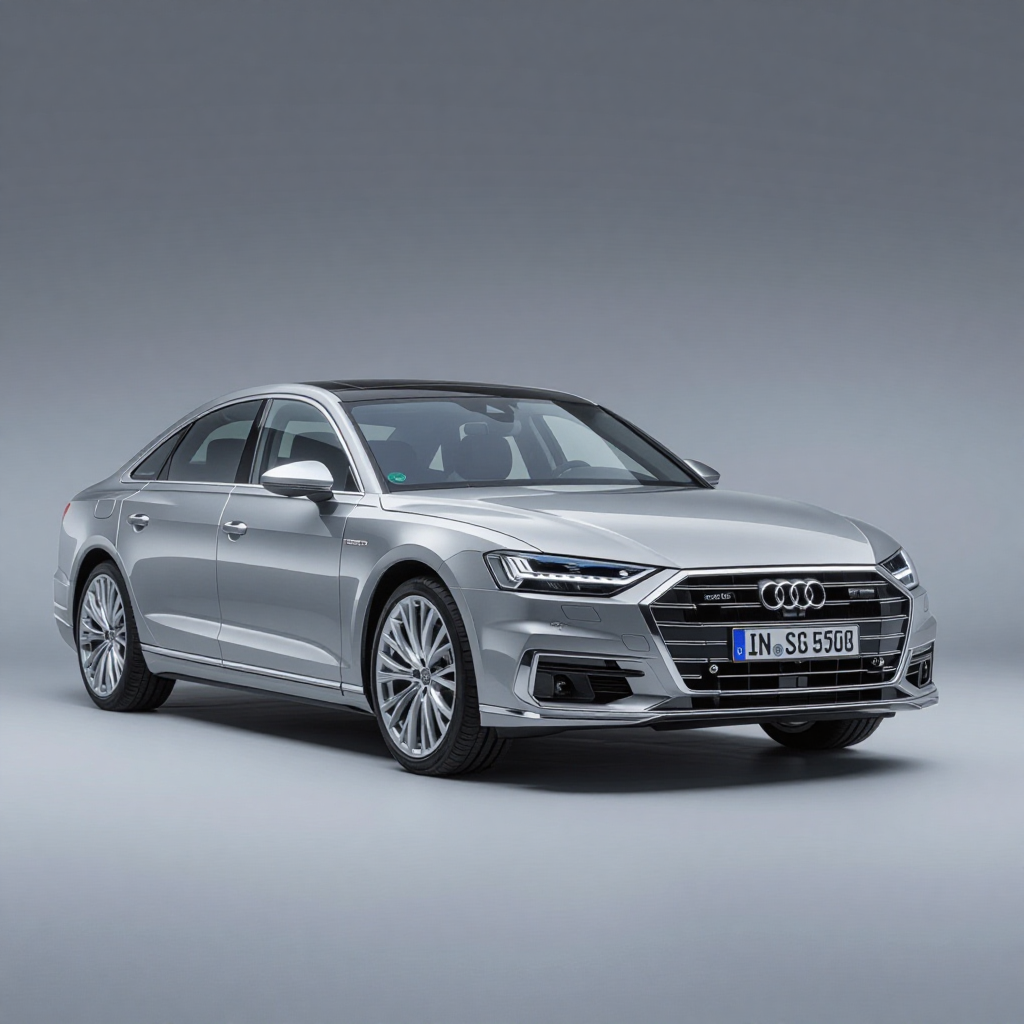}
        \caption{Structurally complete front-oblique view.}
        \label{fig:car_good1}
    \end{subfigure}
    \hfill
    \begin{subfigure}[t]{0.24\textwidth}
        \centering
        \includegraphics[width=\linewidth]{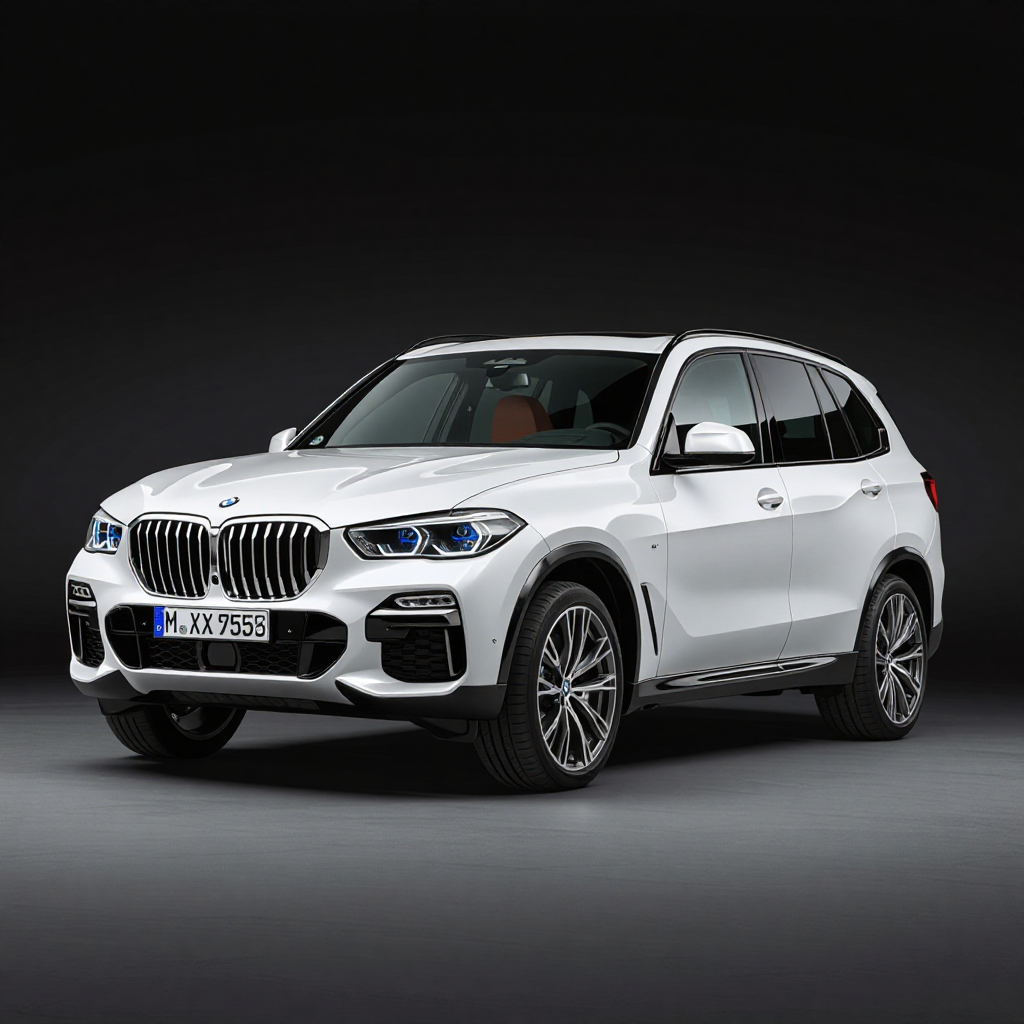}
        \caption{Physically coherent with all components.}
        \label{fig:car_good2}
    \end{subfigure}
    \hfill
    \begin{subfigure}[t]{0.24\textwidth}
        \centering
        \includegraphics[width=\linewidth]{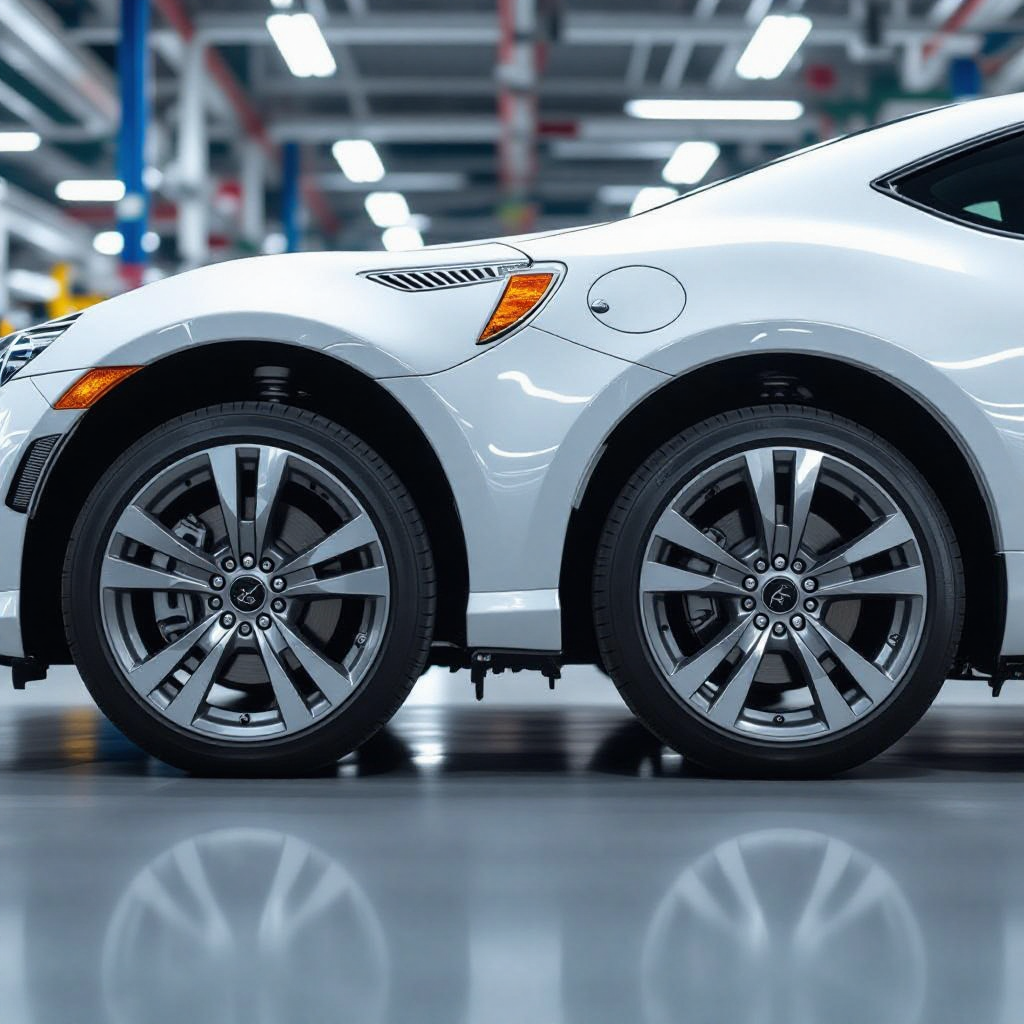}
        \caption{Incorrect wheel placement and missing doors.}
        \label{fig:car_bad1}
    \end{subfigure}
    \hfill
    \begin{subfigure}[t]{0.24\textwidth}
        \centering
        \includegraphics[width=\linewidth]{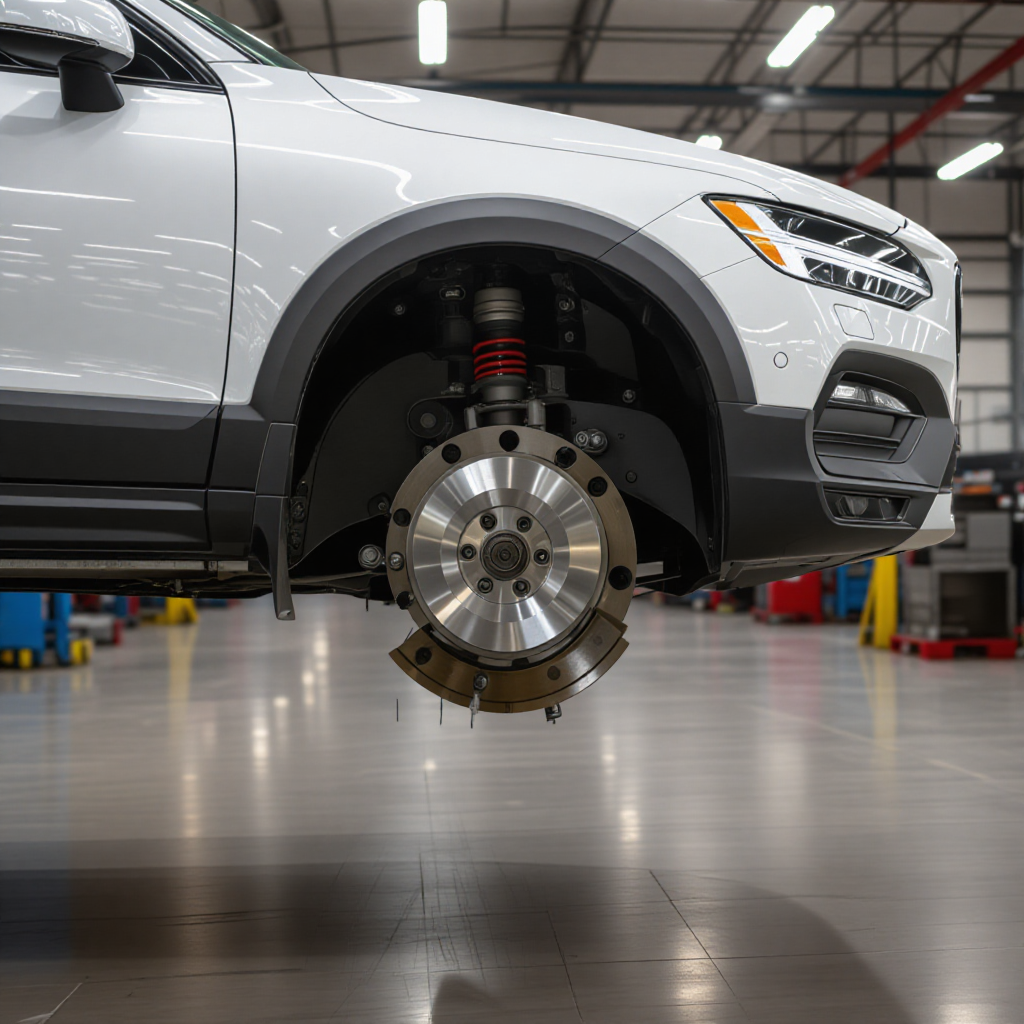}
        \caption{Missing wheels and bumpers; floating vehicle.}
        \label{fig:car_bad2}
    \end{subfigure}

    \caption{Illustrations of structurally sound and physically impossible car.}
    \label{fig:sample_cars}
\end{figure}

\begin{table}[H]
    \centering
    \scriptsize
    \caption{\textbf{Metric scores on car images with varying structural completeness}}
    \label{tab:car_diagnostic}
    \begin{tabularx}{\textwidth}{lccccX}
        \toprule
        \textbf{Image} & \textbf{PCMDE} & \textbf{CLIP} & \textbf{VQA} & \textbf{SigLip-2} & \textbf{Observed Structural Characteristics} \\
        \midrule
        \textbf{(a)} & 80.6 & 29.49 & 0.85 & 0.99 &
        Audi S8 with all critical components. \\
        \textbf{(b)} & 91.6 & 15.76 & 0.28 & 0.99 &
        BMW X5 SUV with complete component set \\
        \textbf{(c)} & 48.0 & 24.07 & 0.70 & 0.99 &
        Side view; missing doors and unreal wheel placement. \\
        \textbf{(d)} & 46.0 & 24.10 & 0.78 & 0.99 &
        Side view lack grille, wheels, bumpers; vehicle lifted above ground. \\
        \bottomrule
    \end{tabularx}
\end{table}

\begin{table}[H]
\centering
\caption{PCMDE vs. Model-as-Judge baselines for car images}
\label{tab:car_ablation}
\small
\begin{tabular}{lccccc}
\toprule
\textbf{Image} & \textbf{PCMDE} & \textbf{LLM-only} & \textbf{VLM-only} & \textbf{YOLO-only} & \textbf{Rules-only} \\
\midrule
(a) & 80.6 & 70.0 & 50.0 & 45.0 & 84.4 \\
(b) & 91.6 & 85.0 & 85.0 & 85.0 & 89.3 \\
(c) & 48.0 & 45.0 & 85.0 & 45.0 & 28.7 \\
(d) & 46.0 & 45.0 & 45.0 & 45.0 & 50.0 \\
\bottomrule
\end{tabular}
\end{table}

\noindent\textbf{Interpretation of Results:} As illustrated in Table~\ref{tab:car_ablation}, cases (a) and (b) are structurally coherent car images and receive high PCMDE scores of 80.6 and 91.6, respectively. Case (a) also shows why detector-only or VLM-only judging is incomplete: VLM-only and YOLO-only assign lower scores of 50.0 and 45.0 despite the image being structurally plausible, while Rules-only remains high at 84.4. Case (b) is consistently scored as valid by the individual variants and by the full pipeline, with PCMDE at 91.6 and Rules-only at 89.3.

Case (c) contains missing doors and unrealistic wheel placement. VLM-only assigns a high score of 85.0, but Rules-only penalizes the spatial violation at 28.7, and the full PCMDE score is 48.0, below the pass threshold. Case (d) is also flagged as implausible, with a PCMDE score of 46.0, due to missing wheels, missing bumpers, and a vehicle body that appears unsupported. These examples show that PCMDE reduces both false acceptance from semantic-only judges and false rejection from detector-only evidence by combining component detections with deterministic structural rules.
\begin{figure}[H]
    \centering

    \begin{subfigure}[t]{0.24\textwidth}
        \centering
        \includegraphics[width=\linewidth]{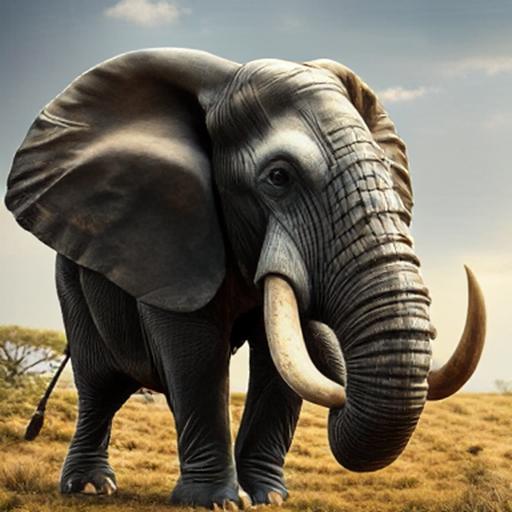}
        \caption{ Proportion failure: oversized head relative to torso.}
        \label{fig:animal_good1}
    \end{subfigure}
    \hfill
    \begin{subfigure}[t]{0.24\textwidth}
        \centering
        \includegraphics[width=\linewidth]{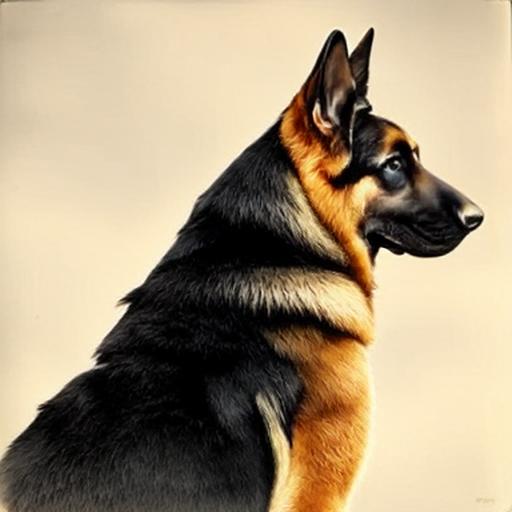}
        \caption{Physically plausible configuration.}
        \label{fig:animal_good2}
    \end{subfigure}
    \hfill
    \begin{subfigure}[t]{0.24\textwidth}
        \centering
        \includegraphics[width=\linewidth]{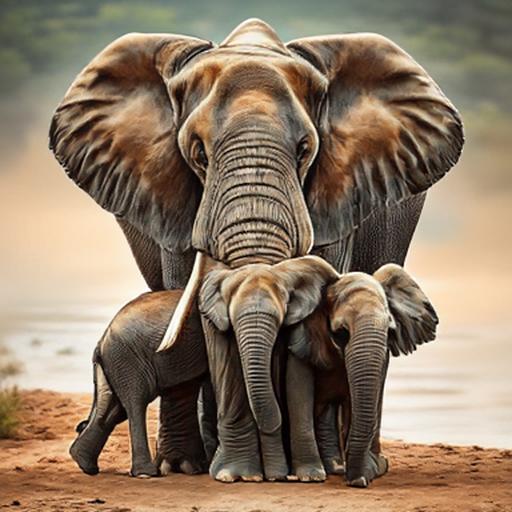}
        \caption{Anatomical merging across subjects.}
        \label{fig:animal_bad1}
    \end{subfigure}
    \hfill
    \begin{subfigure}[t]{0.24\textwidth}
        \centering
        \includegraphics[width=\linewidth]{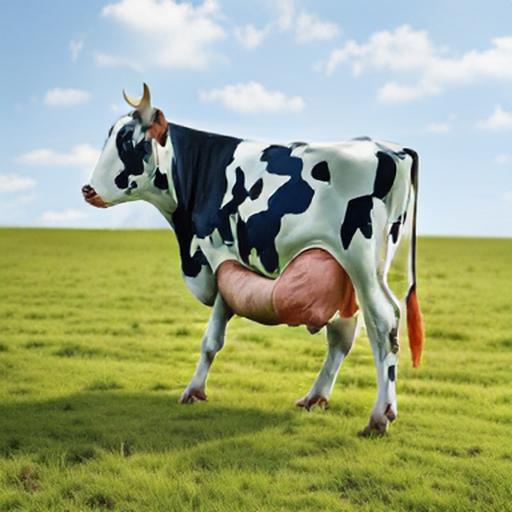}
        \caption{Distorted proportions and leg anomalies.}
        \label{fig:animal_bad2}
    \end{subfigure}

    \caption{Illustrations of structurally sound and physically impossible animal.}
    \label{fig:sample_animal}
\end{figure}
\begin{table}[H]
    \centering
    \scriptsize
    \caption{\textbf{Metric scores on animal images with varying physical plausibility}}
    \label{tab:animal_diagnostic}
    \begin{tabularx}{\textwidth}{lccccX}
        \toprule
        \textbf{Image} & \textbf{PCMDE} & \textbf{CLIP} & \textbf{VQA} & \textbf{SigLip-2} & \textbf{Observed Physical Characteristics} \\
        \midrule
        \textbf{(a)} & 55.4 & 31.6 & 79.8 & 99.9 & Failed due to major fault in head-to-torso proportions (excessively large head); remaining anatomical features are correctly positioned. \\
        \textbf{(b)} & 87.6 & 36.2 & 74.8 & 100.0 & High-quality generation: Structurally consistent canine model with no spatial or component violations. \\
        \textbf{(c)} & 42.2 & 30.3 & 79.8 & 100.0 & Critical anatomical failure: Mixed-up subjects where the adult elephant's trunk incorrectly merges into its offspring. \\
        \textbf{(d)} & 51.8 & 29.0 & 74.9 & 100.0 & Critical failure: Features severe anatomical distortion with a missing leg. \\
        \bottomrule
    \end{tabularx}
\end{table}

\begin{table}[H]
\centering
\caption{PCMDE vs. Model-as-Judge baselines for animal images}
\label{tab:animal_comparison}
\scriptsize
\begin{tabular}{lccccc}
\toprule
\textbf{Image} & \textbf{PCMDE} & \textbf{LLM-only} & \textbf{VLM-only} & \textbf{Zero-Shot-only} & \textbf{Rules-only} \\
\midrule
(a) & 55.4 & 72.0 & 75.0 & 45.0 & 35.0 \\
(b) & 87.6 & 79.0 & 83.0 & 65.0 & 92.0 \\
(c) & 42.2 & 48.0 & 65.0 & 40.0 & 25.0 \\
(d) & 51.8 & 70.0 & 75.0 & 50.0 & 33.0 \\
\bottomrule
\end{tabular}
\end{table}

\noindent\textbf{Interpretation of Results:} Table~\ref{tab:animal_comparison} shows the zoological domain evaluation, which demonstrates the ability of our framework to detect anatomical failures that are missed by standard metrics. Case (b) is a well-structured and high-quality dog that passed all validation stages. Here, we see that Rules-only (92.0) and Full-Pipeline (87.6) closely agree. In case (a), head size disproportion was a major factor, and it was specifically caught by the rule-based spatial stage (Rules-only = 35.0), while VLM-only (75.0) and LLM-only (72.0) failed to catch it. Cases (c) and (d) exhibit critical failures: Case (c) pertains to anatomical fusion, wherein an elephant's trunk integrates with its progeny. The VLM-only approach was excessively lenient at 65.0, and the LLM-only method slightly faltered at 48.0, whereas the Rules-only framework accurately adjusted the score to 25.0, culminating in a Full-Pipeline failure score of 42.2. Case (d) illustrates a cow with a missing leg, wherein both VLM-only (75.0) and LLM-only (70.0) overestimated structural integrity, while Rules-only (33.0) accurately recognized the anatomical discrepancies. In all failure cases, the integrated PCMDE was able to detect biological impossibility by rooting semantic reasoning in deterministic anatomical rules, a fix that neither VLM nor LLM evaluators could accomplish on their own.

\subsection{Agreement with Human Annotators}
\label{sec:human_agreement}

To verify that PCMDE reflects human visual judgment of physical plausibility, we compared PCMDE decisions against majority-vote annotations from three independent human annotators. Human annotation protocol. Annotators were shown each generated image together with its caption and asked to label whether the visible structure was physically plausible. They were instructed to consider component counts, spatial placement, viewpoint-dependent visibility, and obvious structural impossibilities. No personal or sensitive data were collected from annotators, and annotations were used only in aggregate for threshold selection and validation. Annotators were asked to label each synthetic image as either physically plausible or physically implausible based on visible component counts, spatial placement, and structural consistency. For each image \(i\), the human reference label was computed as
\[
y_i^{\mathrm{human}} =
\mathrm{majority}(a_{i,1}, a_{i,2}, a_{i,3}),
\]
where \(a_{i,j}\) is the binary label assigned by annotator \(j\). PCMDE was converted into a binary decision using the same dual-threshold rule used in the main pipeline:
\[
y_i^{\mathrm{PCMDE}} =
\mathbb{1}
\left[
S_i \geq \tau
\;\wedge\;
S_{i,\mathrm{rules}} \geq \tau_c
\;\wedge\;
S_{i,\mathrm{LLM}} \geq \tau_c
\right],
\]
with \(\tau = 60\) and \(\tau_c = 40\). We define human agreement as the fraction of samples for which the PCMDE verdict matches the majority human label:
\[
\mathrm{Agreement}
=
\frac{1}{N}
\sum_{i=1}^{N}
\mathbb{1}
\left[
y_i^{\mathrm{PCMDE}} =
y_i^{\mathrm{human}}
\right].
\]

Table~\ref{tab:human_agreement} shows that PCMDE achieves strong agreement with human visual judgment across all three domains. Overall, PCMDE matches the majority human annotation in \(200/210\) images, corresponding to \(95.2\%\) agreement. This indicates that the proposed score is not merely more variable than embedding-based metrics, but is also aligned with human assessment of physical and structural plausibility.

\begin{table}[H]
\centering
\scriptsize
\caption{Agreement between PCMDE and majority-vote human annotations. A match is counted when the PCMDE PASS/FAIL verdict agrees with the majority label from three human annotators.}
\label{tab:human_agreement}
\begin{tabular}{lcccc}
\toprule
Domain & Images & Annotators & PCMDE--Human Matches & Agreement \\
\midrule
Aircraft & 70 & 3 & 67 / 70 & 95.7\% \\
Car      & 70 & 3 & 66 / 70 & 94.3\% \\
Animal   & 70 & 3 & 67 / 70 & 95.7\% \\
\midrule
Overall  & 210 & 3 & 200 / 210 & 95.2\% \\
\bottomrule
\end{tabular}
\end{table}

\subsection{Case-Study Separation Between Plausible and Implausible Images}
\label{sec:case_study_separation}

We further analyze whether each metric separates physically plausible and implausible case-study examples. For each domain, images (a) and (b) are treated as physically plausible, while images (c) and (d) are treated as physically implausible according to the observed structural characteristics in the case-study tables. To make the comparison easier to interpret, VQAScore and SigLIP-2 are rescaled to the 0--100 range.

Table~\ref{tab:case_study_separation} shows that PCMDE produces a large positive separation between plausible and implausible examples. Across the aircraft and car case studies, PCMDE assigns plausible images an average score of \(88.4\), while implausible images receive an average score of \(50.4\), yielding a \(38.0\)-point separation. In contrast, CLIPScore produces only a \(0.1\)-point separation, VQAScore assigns higher average scores to implausible images, and SigLIP-2 remains almost saturated. PCMDE also ranks every plausible--implausible pair correctly across the aircraft and car case studies.

\begin{table}[H]
\centering
\scriptsize
\caption{Case-study separation between physically plausible and implausible examples. VQAScore and SigLIP-2 are rescaled to 0--100. Pairwise accuracy is the fraction of plausible--implausible image pairs where the metric assigns a higher score to the plausible image; ties count as 0.5.}
\label{tab:case_study_separation}
\begin{tabular}{llcccc}
\toprule
Domain & Metric & Plausible Mean & Implausible Mean & Separation & Pairwise Acc. \\
\midrule
Aircraft & PCMDE    & 90.7  & 53.8  & +36.9 & 100.0\% \\
Aircraft & CLIP     & 31.9  & 30.3  & +1.6  & 100.0\% \\
Aircraft & VQA      & 80.5  & 83.5  & -3.0  & 25.0\% \\
Aircraft & SigLIP-2 & 100.0 & 99.5  & +0.5  & 75.0\% \\
\midrule
Car & PCMDE    & 86.1 & 47.0 & +39.1 & 100.0\% \\
Car & CLIP     & 22.6 & 24.1 & -1.5  & 50.0\% \\
Car & VQA      & 56.5 & 74.0 & -17.5 & 50.0\% \\
Car & SigLIP-2 & 99.0 & 99.0 & 0.0   & 50.0\% \\
\midrule
Aircraft + Car & PCMDE    & 88.4 & 50.4 & +38.0 & 100.0\% \\
Aircraft + Car & CLIP     & 27.3 & 27.2 & +0.1  & 62.5\% \\
Aircraft + Car & VQA      & 68.5 & 78.8 & -10.3 & 43.8\% \\
Aircraft + Car & SigLIP-2 & 99.5 & 99.3 & +0.3  & 62.5\% \\
\bottomrule
\end{tabular}
\end{table}

These case studies support the central design choice of PCMDE. The images remain semantically aligned with their captions, so embedding-based and VQA-based metrics often assign high scores even when the image violates physical structure. PCMDE separates these cases because it grounds the evaluation in detected components, physical rules, and specification-aware reasoning. This demonstrates that the hybrid design provides information that is not captured by semantic similarity alone.

\subsection{Aircraft Rules} Aircraft physical rules are derived based on standard aircraft structural consistency.
This appendix provides the complete specification of physical rules for aircraft images, following the same structure as the aircraft rules described in Section 3.2 of the main paper.

\subsubsection{Presence Rules}

These rules verify that mandatory components exist in realistic quantities.

\begin{compactitem}
    \item \textbf{P1}: Exactly one head must be present: $N_{\text{head}} = 1$
    \item \textbf{P2}: Exactly one tail must be present: $N_{\text{tail}} = 1$
    \item \textbf{P3}: Engine count must be 2, 3, or 4 for commercial aircraft: $N_{\text{engine}} \in \{2, 3, 4\}$
    \item \textbf{P4}: One or two swept wings must be visible (depending on viewpoint): $N_{\text{wing}} \in \{1, 2\}$
    \item \textbf{P5}: Engine width must be reasonable relative to fuselage: $0.05 \cdot L_{\text{fuselage}} < W_{\text{engine}} < 0.15 \cdot L_{\text{fuselage}}$
\end{compactitem}

\subsubsection{Spatial Rules}

These rules enforce geometric relationships between component bounding boxes.

\begin{compactitem}
    \item \textbf{S1}: Engines must be positioned near wings. For each engine bounding box $b_e$, there exists a wing bounding box $b_w$ such that:
    \begin{itemize}
        \item Either: $d_y(b_e, b_w) < 50$ pixels (vertical proximity), where $d_y(b_i, b_j) = |\text{centroid}_y(b_i) - \text{centroid}_y(b_j)|$
        \item Or: the centroid of $b_e$ lies within the expanded wing region $b_w^+ = [x_1 - 30, y_1 - 80, x_2 + 30, y_2 + 80]$
    \end{itemize}
    
    \item \textbf{S2}: Head must be at the front (left side for side-view aircraft). This enforces: $\text{centroid}_x(b_{\text{head}}) < \text{centroid}_x(b_{\text{tail}})$, where $\text{centroid}_x(b) = (x_1 + x_2) / 2$
    
    \item \textbf{S3}: Tail must be at the rear of the fuselage: $\text{centroid}_x(b_{\text{tail}}) > 0.7 \cdot W$, where $W$ is the image width
    
    \item \textbf{S4}: Wings must intersect the fuselage centerline (at $y = H/2$, where $H$ is image height)
\end{compactitem}

\subsubsection{Relational Rules}

These rules encode logical dependencies.

\begin{compactitem}
    \item \textbf{R1}: If engines are detected, wings must also be present: $N_{\text{engine}} > 0 \Rightarrow N_{\text{wing}} > 0$
    \item \textbf{R2}: Head and tail must exist together: $N_{\text{head}} > 0 \Leftrightarrow N_{\text{tail}} > 0$
    \item \textbf{R3}: Component size ratios must be consistent with real-world proportions: $W_{\text{engine}} \propto L_{\text{fuselage}}$ with ratio in range $[0.05, 0.15]$
\end{compactitem}

\subsubsection{Engine Placement Variants by Aircraft Type}

Different aircraft types have distinct engine configurations:

\begin{compactitem}
    \item \textbf{Under-Wing}: Engines below wings ($\text{centroid}_y(b_e) > \text{centroid}_y(b_w)$). Common in commercial airliners (Boeing 737, 747, Airbus A320, A380).
    
    \item \textbf{Rear-Fuselage}: Engines on rear fuselage sides ($x_{\text{center}}(b_e) > x_{\text{max}}(b_{\text{fuselage}}) - \text{margin}$). Common in business jets and some regional jets.
    
    \item \textbf{Tail-Mounted}: Central engine at tail base ($x_{\text{center}}(b_e) \approx x_{\text{center}}(b_{\text{tail}})$). Used in tri-jet configurations (DC-10, MD-11, Boeing 727).
\end{compactitem}

\subsubsection{View-Specific Constraints}

Component visibility varies by viewpoint:

\begin{table}[H]
\centering
\scriptsize
\begin{tabular}{|l|c|c|}
\hline
\textbf{Component} & \textbf{Side View} & \textbf{Front/Rear View} \\
\hline
Head & 1 & 1 \\
Tail & 1 & 1 \\
Swept Wings & 1 & 2 \\
Engines (twin) & 1--2 & 2 \\
Engines (tri-jet) & 2--3 & 3 \\
Engines (quad) & 2--4 & 4 \\
\hline
\end{tabular}
\caption{Expected component counts by viewpoint. For side views, only visible engines are counted.}
\label{tab:aircraft_view_constraints}
\end{table}

\subsubsection{Rule Evaluation}

Rules are evaluated using the scoring framework from Section 3.2:

\begin{equation}
s_{\text{pres}} = \frac{\text{Satisfied presence rules (P1--P5)}}{\text{Total applicable presence rules}}
\end{equation}

\begin{equation}
s_{\text{spat}} = \frac{\text{Satisfied spatial rules (S1--S4)}}{\text{Total applicable spatial rules}}
\end{equation}

\begin{equation}
s_{\text{rel}} = \frac{\text{Satisfied relational rules (R1--R3)}}{\text{Total applicable relational rules}}
\end{equation}

Final rule-based score: $S_{\text{rules}}^{\text{aircraft}} = 100 \cdot (0.35 \cdot s_{\text{pres}} + 0.25 \cdot s_{\text{spat}} + 0.25 \cdot s_{\text{rel}} + 0.15 \cdot s_{\text{cap}})$

where $s_{\text{cap}}$ is the caption alignment score (Section 3.2.4).

\subsection{Car Rules}
This appendix provides the complete specification of physical rules for car images, following the same structure as the aircraft rules described in Section 3.2 of the main paper.

\subsubsection{Presence Rules}

These rules verify that mandatory components exist in realistic quantities, accounting for viewpoint-dependent visibility.

\begin{compactitem}
    \item \textbf{P1}: Wheel count must be consistent with vehicle type and viewpoint: $N_{\text{wheels}} \in \{2, 3, 4\}$ (4 total inferred, 2--3 visible in side view, 2--4 in top/oblique view)
    
    \item \textbf{P2}: Exactly one bonnet, windshield, and roof when visible: $N_{\text{bonnet}} = N_{\text{windshield}} = N_{\text{roof}} = 1$
    
    \item \textbf{P3}: Headlights and taillights must appear in even pairs: $N_{\text{headlights}}, N_{\text{taillights}} \in \{0, 2\}$ (view-dependent: 2 in front/rear view, 0--2 in side view)
    
    \item \textbf{P4}: Door count must match vehicle type: $N_{\text{doors}} \in \{2, 4\}$ (sedan/SUV: 4 doors, coupe: 2 doors; side view shows 1--2, top view shows all)
    
    \item \textbf{P5}: Fender count must match wheel count: $N_{\text{fenders}} = N_{\text{wheels}} = 4$ (one fender per wheel)
    
    \item \textbf{P6}: Side windows must match door count: $N_{\text{windows}} = N_{\text{doors}}$
    
    \item \textbf{P7}: Exactly one grille in front-facing views: $N_{\text{grille}} = 1$ (must be 0 in rear view)
    
    \item \textbf{P8}: Exactly one trunk in rear-facing views: $N_{\text{trunk}} = 1$ (must be 0 in front view)
    
    \item \textbf{P9}: Rearview mirrors in pairs: $N_{\text{mirrors}} \in \{1, 2\}$ (2 expected in side/oblique views)
    
    \item \textbf{P10}: Exactly one front and one rear bumper: $N_{\text{front\_bumper}} = N_{\text{rear\_bumper}} = 1$
\end{compactitem}

\subsubsection{Spatial Rules}

These rules enforce geometric relationships between component bounding boxes, defined in normalized image coordinates.

\begin{compactitem}
    \item \textbf{S1}: Wheels must be at the bottom of the vehicle: $\text{centroid}_y(b_{\text{wheel}}) > 0.6 \cdot H$ for each wheel (ground contact requirement)
    
    \item \textbf{S2}: Roof must be at the top: $\text{centroid}_y(b_{\text{roof}}) < 0.4 \cdot H$
    
    \item \textbf{S3}: Bonnet at front section: $\text{centroid}_x(b_{\text{bonnet}}) < 0.4 \cdot W$ (for side view facing right)
    
    \item \textbf{S4}: Trunk at rear section: $\text{centroid}_x(b_{\text{trunk}}) > 0.6 \cdot W$ (for side view facing right)
    
    \item \textbf{S5}: Windshield positioned between bonnet and roof: $x_{\text{max}}(b_{\text{bonnet}}) \leq x_{\text{min}}(b_{\text{windshield}}) \leq x_{\text{max}}(b_{\text{roof}})$ and $y_{\text{max}}(b_{\text{windshield}}) < y_{\text{min}}(b_{\text{roof}})$
    
    \item \textbf{S6}: Headlights at front corners: $\text{centroid}_x(b_{\text{headlight}}) < 0.3 \cdot W$ (symmetrically positioned)
    
    \item \textbf{S7}: Taillights at rear corners: $\text{centroid}_x(b_{\text{taillight}}) > 0.7 \cdot W$ (symmetrically positioned)
    
    \item \textbf{S8}: Doors between front and rear wheels: $x_{\text{min}}(b_{\text{wheel}_{\text{front}}}) < \text{centroid}_x(b_{\text{door}}) < x_{\text{max}}(b_{\text{wheel}_{\text{rear}}})$
    
    \item \textbf{S9}: Grille below bonnet at front: $y_{\text{center}}(b_{\text{grille}}) > y_{\text{center}}(b_{\text{bonnet}})$ and $x_{\text{center}}(b_{\text{grille}}) \approx x_{\text{center}}(b_{\text{bonnet}})$
    
    \item \textbf{S10}: Side windows directly above doors: $x_{\text{overlap}}(b_{\text{door}}, b_{\text{window}}) > 0.7 \cdot \text{width}(b_{\text{door}})$ and $y_{\text{min}}(b_{\text{window}}) < y_{\text{max}}(b_{\text{door}})$
    
    \item \textbf{S11}: Fenders positioned around wheels: Each fender $f_i$ must satisfy $\text{IoU}(b_{f_i}, b_{w_i}^+) > 0.3$, where $b_{w_i}^+$ is expanded wheel region
    
    \item \textbf{S12}: Headlights and taillights must be on opposite ends: $|x_{\text{center}}(b_{\text{headlight}}) - x_{\text{center}}(b_{\text{taillight}})| > 0.5 \cdot W$
\end{compactitem}

\subsubsection{Relational Rules}

These rules encode logical dependencies between components.

\begin{compactitem}
    \item \textbf{R1}: If wheels are detected, a vehicle body must exist: $N_{\text{wheels}} > 0 \Rightarrow (N_{\text{bonnet}} > 0 \text{ OR } N_{\text{trunk}} > 0 \text{ OR } N_{\text{roof}} > 0)$
    
    \item \textbf{R2}: Each wheel must be enclosed by a fender: $\forall$ wheel $w_i$, $\exists$ fender $f_i$ such that $\text{bbox}(w_i) \subset \text{bbox}(f_i)^+$
    
    \item \textbf{R3}: Headlights and taillights must exist together on complete vehicles: $(N_{\text{headlights}} > 0) \Leftrightarrow (N_{\text{taillights}} > 0)$ (when both views visible)
    
    \item \textbf{R4}: Doors must not intersect wheel regions: $\text{IoU}(b_{\text{door}}, b_{\text{wheel}}) = 0$ for all door-wheel pairs
    
    \item \textbf{R5}: Windshield connects bonnet to roof: $y_{\text{max}}(b_{\text{windshield}}) \approx y_{\text{min}}(b_{\text{roof}})$ and $y_{\text{min}}(b_{\text{windshield}}) \approx y_{\text{max}}(b_{\text{bonnet}})$
    
    \item \textbf{R6}: Component size ratios must be realistic:
    \begin{itemize}
        \item Wheel diameter: $0.10 \leq D_{\text{wheel}} / H_{\text{vehicle}} \leq 0.25$
        \item Side window area: $\text{area}(b_{\text{window}}) < 0.7 \cdot \text{area}(b_{\text{door}})$
        \item Headlight/taillight size: $\text{area}(b_{\text{light}}) < 0.15 \cdot \text{area}(b_{\text{bonnet/trunk}})$
    \end{itemize}
    
    \item \textbf{R7}: Roof must span above all doors: $x_{\text{min}}(b_{\text{roof}}) \leq x_{\text{min}}(\text{all doors})$ and $x_{\text{max}}(b_{\text{roof}}) \geq x_{\text{max}}(\text{all doors})$
    
    \item \textbf{R8}: Grille must be adjacent to headlights horizontally: $|x_{\text{center}}(b_{\text{grille}}) - x_{\text{center}}(b_{\text{headlight}_i})| < 0.3 \cdot W$ for each headlight in front view
    
    \item \textbf{R9}: Symmetry constraint for paired components: For headlights, taillights, and mirrors, left and right components must be approximately symmetric: $|y_{\text{center}}(b_L) - y_{\text{center}}(b_R)| < 0.1 \cdot H$
    
    \item \textbf{R10}: Bumpers align with lights: Front bumper below headlights ($y_{\text{min}}(b_{\text{front\_bumper}}) \approx y_{\text{max}}(b_{\text{headlights}})$), rear bumper below taillights
\end{compactitem}

\subsubsection{View-Specific Constraints}

Component visibility and count expectations vary by viewpoint:

\begin{table}[H]
\centering
\scriptsize
\begin{tabular}{|l|c|c|c|c|}
\hline
\textbf{Component} & \textbf{Front View} & \textbf{Rear View} & \textbf{Side View} & \textbf{Top View} \\
\hline
Wheels & 2 & 2 & 2--3 & 4 \\
Headlights & 2 & 0 & 1--2 & 0 \\
Taillights & 0 & 2 & 1--2 & 0 \\
Doors & 0 & 0 & 1--2 & 2--4 \\
Bonnet & 1 & 0 & 1 & 1 \\
Trunk & 0 & 1 & 1 & 1 \\
Grille & 1 & 0 & 0--1 & 0 \\
Windshield & 1 & 0 & 1 & 1 \\
Roof & 0--1 & 0--1 & 1 & 1 \\
Mirrors & 0--1 & 0--1 & 2 & 0--2 \\
\hline
\end{tabular}
\caption{Expected component counts by viewpoint. 0 indicates component should not be visible.}
\label{tab:car_view_constraints}
\end{table}

\subsubsection{Vehicle Type-Specific Rules}

Additional constraints apply based on vehicle type mentioned in captions:

\begin{compactitem}
    \item \textbf{Sedan}: 4 doors, 4 wheels, conventional trunk, standard roof height
    \item \textbf{Coupe}: 2 doors, 4 wheels, sloped roofline, integrated trunk
    \item \textbf{SUV/Truck}: 4 doors, 4--6 wheels (trucks may have dual rear wheels), elevated body position ($y_{\text{min}}(\text{body}) > 0.5 \cdot H$)
    \item \textbf{Hatchback}: 2--4 doors, 4 wheels, integrated trunk/roof (no distinct trunk separation)
\end{compactitem}

\subsubsection{Rule Evaluation}

Rules are evaluated using the scoring framework from Section 3.2:

\begin{equation}
s_{\text{pres}} = \frac{\text{Satisfied presence rules (P1--P10)}}{\text{Total applicable presence rules}}
\end{equation}

\begin{equation}
s_{\text{spat}} = \frac{\text{Satisfied spatial rules (S1--S12)}}{\text{Total applicable spatial rules}}
\end{equation}

\begin{equation}
s_{\text{rel}} = \frac{\text{Satisfied relational rules (R1--R10)}}{\text{Total applicable relational rules}}
\end{equation}

Final rule-based score: $S_{\text{rules}}^{\text{car}} = 100 \cdot (0.35 \cdot s_{\text{pres}} + 0.25 \cdot s_{\text{spat}} + 0.25 \cdot s_{\text{rel}} + 0.15 \cdot s_{\text{cap}})$

where $s_{\text{cap}}$ is the caption alignment score (Section 3.2.4).

\subsection{Animal Rules}
Animal physical rules are derived from standard zoological anatomy and species-specific morphology. This appendix provides the complete specification of physical rules for animal images. Rules are grouped into presence, spatial, relational, and subtype-specific categories. Eight animal subtypes are supported: dog, cat, horse, sheep, cow, squirrel, elephant, and chicken. A fallback \texttt{generic\_animal} subtype is applied when no specific subtype is identified.

\subsubsection{Presence Rules}

These rules verify that mandatory anatomical components exist in realistic quantities.

\begin{compactitem}
    \item \textbf{P1}: Mandatory core components: $\{\text{torso, head}\} \subseteq \text{detected components}$
    
    \item \textbf{P2}: Exactly one torso: $N_{\text{torso}} = 1$
    
    \item \textbf{P3}: Exactly one head: $N_{\text{head}} = 1$
    
    \item \textbf{P4}: Leg count must match species: $N_{\text{leg}} = 4$ for quadrupeds; $N_{\text{leg}} = 2$ for bipeds (chicken)
    
    \item \textbf{P5}: Eye count within plausible range: $N_{\text{eye}} \in \{1, 2\}$ (viewpoint dependent)
    
    \item \textbf{P6}: Tail count: $N_{\text{tail}} \in \{0, 1\}$
    
    \item \textbf{P7}: Wings restricted to avian subtypes: $N_{\text{wing}} = 0$ for all non-avian subtypes
    
    \item \textbf{P8}: Beak restricted to avian subtypes: $N_{\text{beak}} = 0$ for all non-avian subtypes
    
    \item \textbf{P9}: Exactly two wings for birds: $N_{\text{wing}} = 2$ (applies to subtype: bird)
    
    \item \textbf{P10}: Exactly one beak for birds: $N_{\text{beak}} = 1$ (applies to subtype: bird)
    
    \item \textbf{P11}: No foreign anatomical features: $N_{\text{foreign\_part}} = 0$ for all subtypes. The category $\text{foreign\_part}$ includes any detected feature not in the species' canonical part inventory (e.g., human-like hands or faces, anthropomorphic limbs). Detection The detection stage is to detect foreign parts via a curated per-subtype prompt list. This rule fires deterministically upon any such feature being reported. A violation is a \emph{critical anatomical inconsistency} and invokes the score cap $S_{\text{cap}} = 30$.
    
\end{compactitem}

\subsubsection{Spatial Rules}

These rules enforce geometric relationships between component bounding boxes $b$, ensuring anatomically correct arrangements.

\begin{compactitem}
    \item \textbf{S1}: Eyes must lie within the head: $\text{containment}(b_{\text{eye}}, b_{\text{head}}) \geq 0.70$
    
    \item \textbf{S2}: Beak must lie within the head (chicken): $\text{containment}(b_{\text{beak}}, b_{\text{head}}) \geq 0.60$
    
    \item \textbf{S3}: Legs must not overlap head: $\text{IoU}(b_{\text{leg}}, b_{\text{head}}) \leq 0.50$
    
    \item \textbf{S4}: Wings must not overlap head (bird): $\text{IoU}(b_{\text{wing}}, b_{\text{head}}) \leq 0.30$
    
    \item \textbf{S5}: Beak must be anchored to head, not torso: $\text{IoU}(b_{\text{beak}}, b_{\text{torso}}) \leq 0.30$
    
    \item \textbf{S6}: Tail must not overlap eyes: $\text{IoU}(b_{\text{tail}}, b_{\text{eye}}) \leq 0.30$
    
    \item \textbf{S7}: Legs must be attached to torso within a margin: $b_{\text{leg}} \subset \text{margin}(b_{\text{torso}}, 0.25)$
    
    \item \textbf{S8}: Head at anterior end: $\text{centroid}_x(b_{\text{head}}) \approx \min_x(b_{\text{animal}})$ (assuming left-facing pose)
    
    \item \textbf{S9}: Tail at posterior end: $\text{centroid}_x(b_{\text{tail}})$ must be opposite to $\text{centroid}_x(b_{\text{head}})$
    
    \item \textbf{S10}: Legs attach below torso: $\text{centroid}_y(b_{\text{leg}}) > \text{centroid}_y(b_{\text{torso}})$
    
    \item \textbf{S11}: Beak projects forward from head: $\text{centroid}_x(b_{\text{beak}}) < \text{centroid}_x(b_{\text{head}})$ (for left-facing)
\end{compactitem}

\subsubsection{Relational Rules}

These rules encode logical dependencies and proportional constraints.

\begin{compactitem}
    \item \textbf{R1}: Eye presence implies head: $N_{\text{eye}} > 0 \Rightarrow N_{\text{head}} > 0$
    
    \item \textbf{R2}: Leg presence implies torso: $N_{\text{leg}} > 0 \Rightarrow N_{\text{torso}} > 0$
    
    \item \textbf{R3}: Torso must be the largest component: $\text{area}(b_{\text{head}}) < \text{area}(b_{\text{torso}})$
    
    \item \textbf{R4}: Eye size relative to head: $\text{area}(b_{\text{eye}}) < \text{area}(b_{\text{head}})$
    
    \item \textbf{R5}: Proportional Constraints:
    \begin{compactitem}
        \item Head width: $0.15 \leq W_{\text{head}} / W_{\text{torso}} \leq 0.50$
        \item Leg height: $0.30 \leq H_{\text{leg}} / H_{\text{torso}} \leq 1.50$
        \item Tail length: $0.10 \leq W_{\text{tail}} / W_{\text{torso}} \leq 0.80$
        \item Eye diameter: $0.05 \leq W_{\text{eye}} / W_{\text{head}} \leq 0.15$
        \item Wing spread: $0.50 \leq W_{\text{wing}} / W_{\text{torso}} \leq 1.50$ (birds only)
        \item Beak length: $0.20 \leq W_{\text{beak}} / W_{\text{head}} \leq 0.40$ (birds only)
    \end{compactitem}
\end{compactitem}

\subsubsection{View-Specific Constraints}

Component visibility and expected detection counts vary by viewpoint.

\begin{table}[H]
\centering
\scriptsize
\begin{tabular}{|l|c|c|c|}
\hline
\textbf{Component} & \textbf{Side View} & \textbf{Front View} & \textbf{Rear View} \\
\hline
Torso & 1 & 1 & 1 \\
Head & 1 & 1 & 0--1 \\
Leg (Quadruped) & 2--4 & 2--4 & 2--4 \\
Leg (Bird) & 2 & 2 & 2 \\
Eye & 1 & 2 & 0--1 \\
Tail & 1 & 0--1 & 1 \\
Wing (Bird) & 1--2 & 2 & 1--2 \\
Beak (Bird) & 1 & 1 & 0--1 \\
\hline
\end{tabular}
\caption{Expected animal component counts by viewpoint.}
\label{tab:animal_view_constraints}
\end{table}

\subsubsection{Subtype-Specific Rules}

\begin{itemize}
    \item \textbf{Mammals} (Dog, Cat, etc.): Quadrupedal locomotion ($N_{\text{leg}} = 4$), no wings, no beak.
    \item \textbf{Bird} (Chicken): Bipedal ($N_{\text{leg}} = 2$), mandatory wings ($N_{\text{wing}} = 2$) and beak ($N_{\text{beak}} = 1$).
\end{itemize}

\subsubsection{Rule Evaluation}

The scoring for animal images follows the same weighted framework:

\begin{equation}
s_{\text{pres}} = \frac{\text{Satisfied presence rules (P1--P11)}}{\text{Total applicable presence rules}}
\end{equation}

\begin{equation}
s_{\text{spat}} = \frac{\text{Satisfied spatial rules (S1--S11)}}{\text{Total applicable spatial rules}}
\end{equation}

\begin{equation}
s_{\text{rel}} = \frac{\text{Satisfied relational rules (R1--R5)}}{\text{Total applicable relational rules}}
\end{equation}

Final rule-based score: $S_{\text{rules}}^{\text{animal}} = 100 \cdot (0.35 \cdot s_{\text{pres}} + 0.25 \cdot s_{\text{spat}} + 0.25 \cdot s_{\text{rel}} + 0.15 \cdot s_{\text{cap}})$

\subsection{Results Visualization}

\begin{figure}[H]
\centering

\begin{subfigure}[t]{0.32\linewidth}
    \centering
    \includegraphics[width=\linewidth]{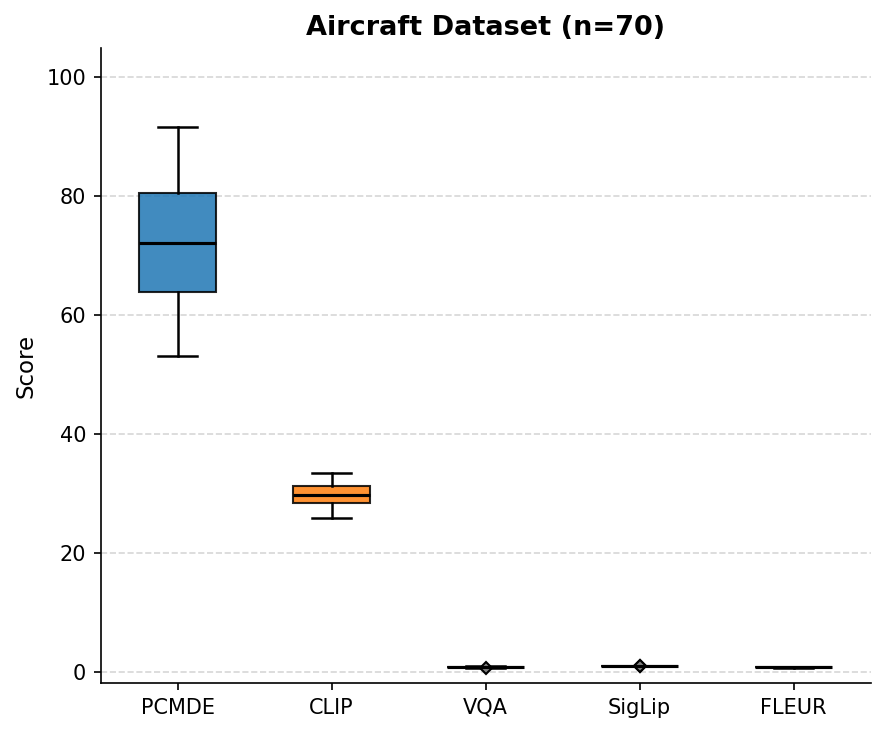}
    \caption{Aircraft (Range=42.5, CV=16.9\%)}
\end{subfigure}
\hfill
\begin{subfigure}[t]{0.32\linewidth}
    \centering
    \includegraphics[width=\linewidth]{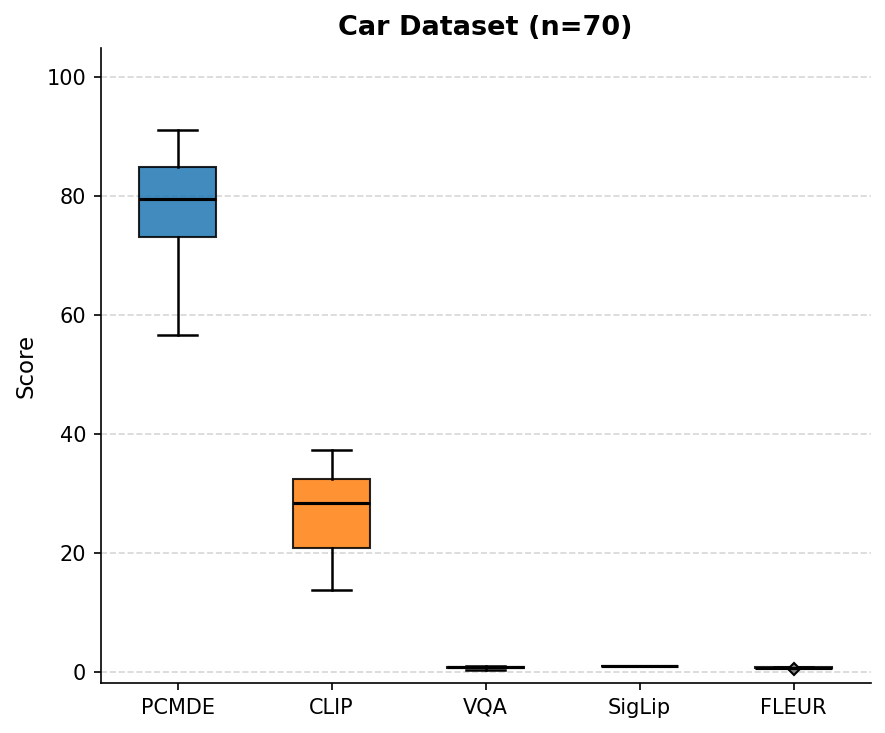}
    \caption{Car (Range=45.6, CV=12.15\%)}
\end{subfigure}
\hfill
\begin{subfigure}[t]{0.32\linewidth}
    \centering
    \includegraphics[width=\linewidth]{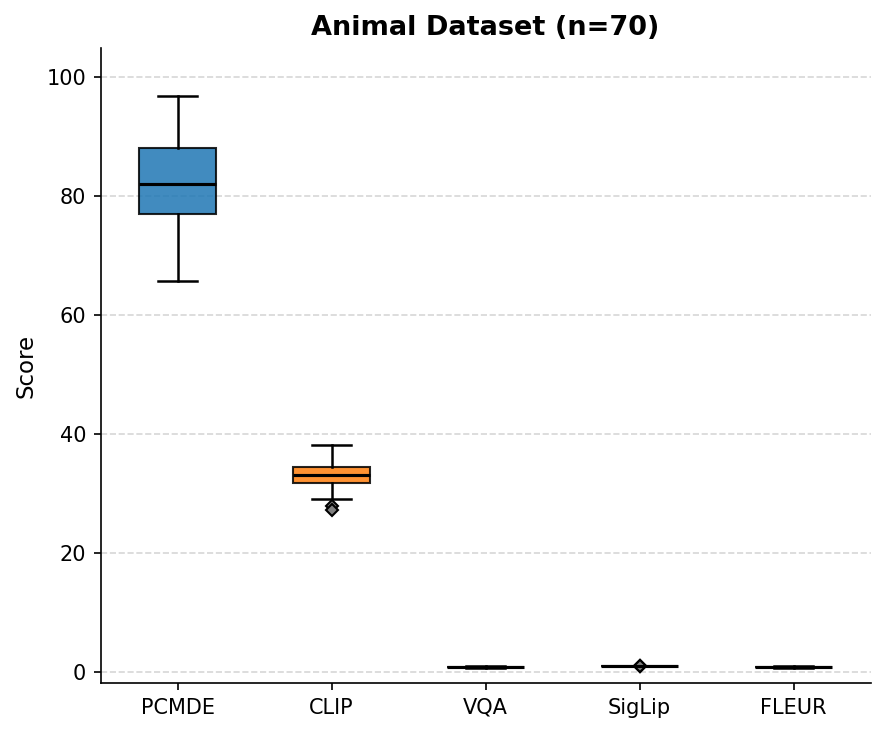}
    \caption{Animal (Range=42.0, CV=10.7\%)}
\end{subfigure}

\caption{\footnotesize Score distributions across 70 synthetic images per domain. PCMDE consistently maintains wide discriminative spread with stable calibration, while SigLIP-2 shows saturation and CLIP/VQA exhibit limited or unstable variability.}
\label{fig:dist_all}
\end{figure}

Figures~\ref{fig:dist_all} show the score distributions for the PCMDE and all baseline metrics for each domain separately. Note that PCMDE and CLIP are reported on a 0-100 scale, whereas VQA, SigLIP-2, and FLEUR are reported on a 0-1 scale. We plot all metrics on a common 0-100 scale to highlight relative discriminative range, and full numerical statistics are presented in Table~\ref{tab:merged_metric_statistics_sidebyside}. PCMDE has the largest absolute score range among all three domains, allowing more fine-grained quality discrimination between physically plausible vs. implausible images. SigLIP-2 is nearly completely saturated in all domains ($\text{CV} \leq 1.5\%$), collapsing to a flat line and unable to distinguish any differences in structure quality. CLIP has a very constrained spread on aircraft and animal domains but broader, unstable variability on cars ($\text{CV} = 22.3\%$) due to extreme low-score samples, not consistent quality differentiation. VQA and FLEUR show moderate statistical variability. They are adjacent to SigLIP-2 on the common axis. This means that the structured, physics-informed separation achieved by PCMDE through its rule-based and LLM reasoning framework cannot be obtained by either the embedding-based or the VQA-based baselines.

\subsection{LLM Prompt Template for Domain-Agnostic Rule Generation}
To guarantee schema compliance and prevent hallucinated constraints, we constrain the LLM with a structured generation prompt. The template enforces a fixed 10-rule catalog, explicit field contracts, severity tiers, and detector-grounded component naming. The complete prompt is provided below:

\begin{figure}[H]
\centering
\begin{minipage}{0.95\linewidth}
\footnotesize
\textbf{Rule Generation Prompt Schema}\\[0.5em]
\texttt{
\{ \\
\quad "object\_type": "<domain>", \\
\quad "subtypes": \{<id>: \{display\_name, filename\_patterns, caption\_regex\}\}, \\
\quad "scoring\_weights": \{"presence": 0.35, "spatial": 0.25, "relational": 0.25, "subtype": 0.15\}, \\
\quad "rules": [ \\
\quad \quad \{ \\
\quad \quad \quad "rule\_type": <one of 10 types>, \\
\quad \quad \quad "subject\_component": "<singular\_lowercase>", \\
\quad \quad \quad "parameters": <type-specific>, \\
\quad \quad \quad "severity": "critical" | "major" | "minor" \\
\quad \quad \} \\
\quad ] \\
\}
}
\\[0.5em]
\textbf{Rule Types:} count\_constraint, containment\_constraint, forbid\_overlap, attachment\_constraint, relative\_position\_constraint, size\_ratio\_constraint, size\_hierarchy, dependency\_constraint, subtype\_restriction, critical\_presence.
\\[0.3em]
\textbf{Key Constraints:} (1) Reference only detector-emitted components; (2) Use singular lowercase nouns; (3) 15–35 rules total; (4) Critical violations cap score at 30.
\end{minipage}
\caption{Structured schema for LLM-generated physical validation rules. The prompt enforces a fixed 10-rule catalog and explicit field contracts to ensure deterministic, machine-executable JSON output.}
\label{fig:rule_schema}
\end{figure}

\subsection{Sample Annotated images from Dataset}

\begin{figure}[htbp]
    \centering

    \begin{subfigure}[t]{0.23\textwidth}
        \centering
        \includegraphics[width=\linewidth]{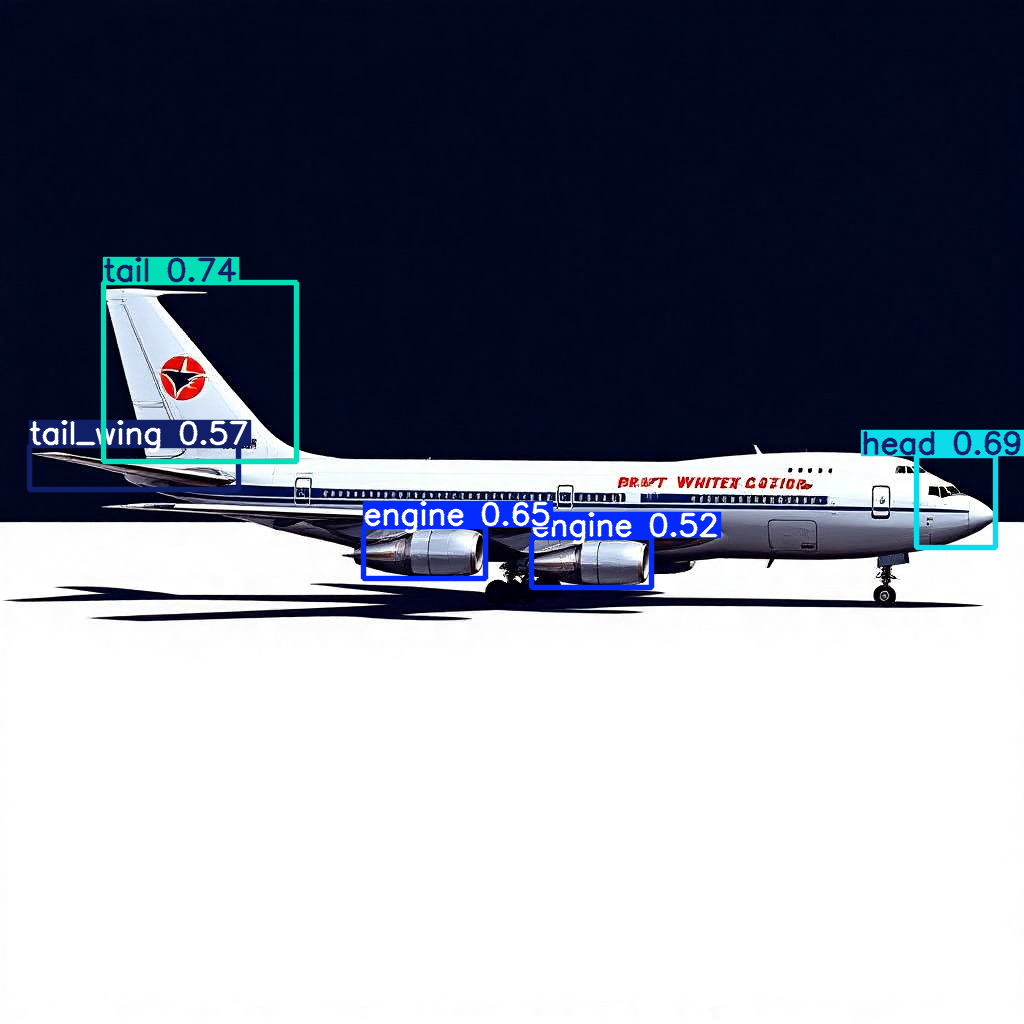}
        \caption{Image 1}
        \label{fig:img1}
    \end{subfigure}
    \hfill
    \begin{subfigure}[t]{0.23\textwidth}
        \centering
        \includegraphics[width=\linewidth]{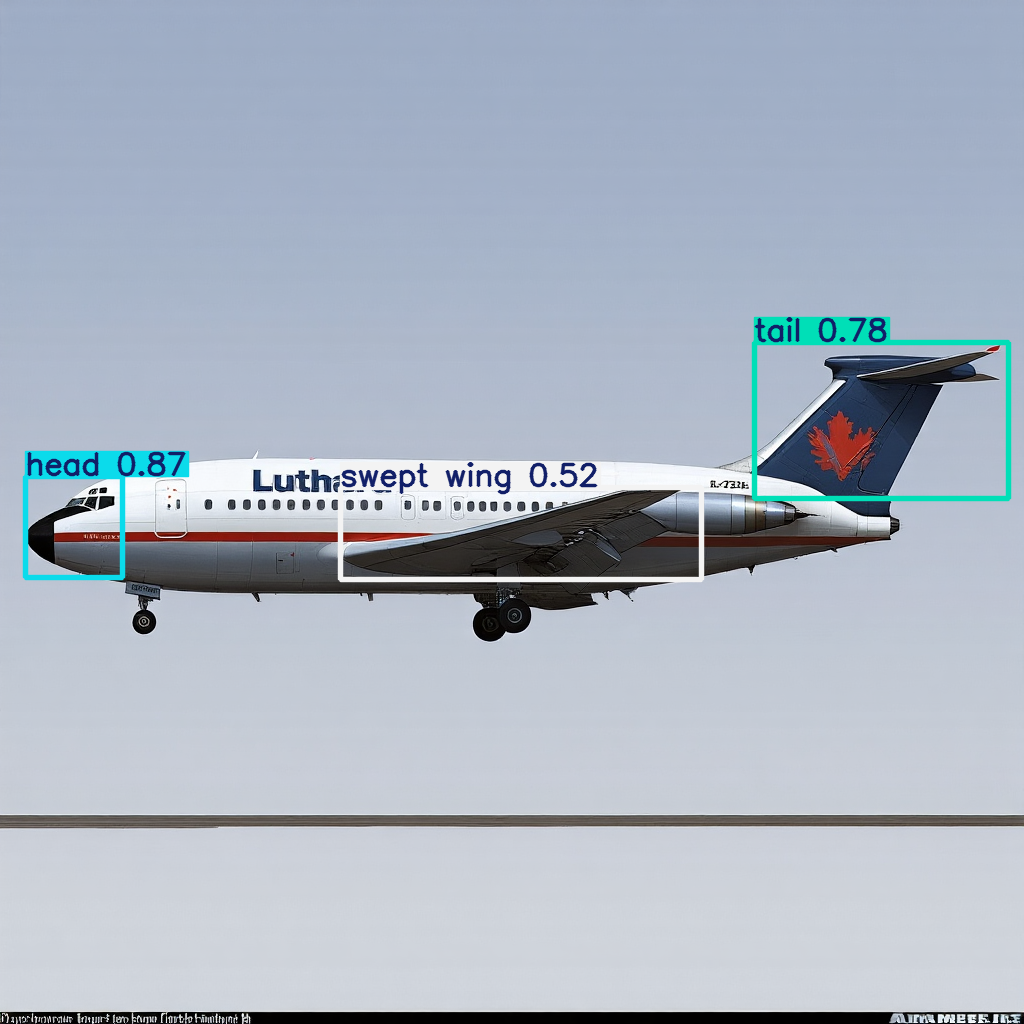}
        \caption{Image 2}
        \label{fig:img2}
    \end{subfigure}
    \hfill
    \begin{subfigure}[t]{0.23\textwidth}
        \centering
        \includegraphics[width=\linewidth]{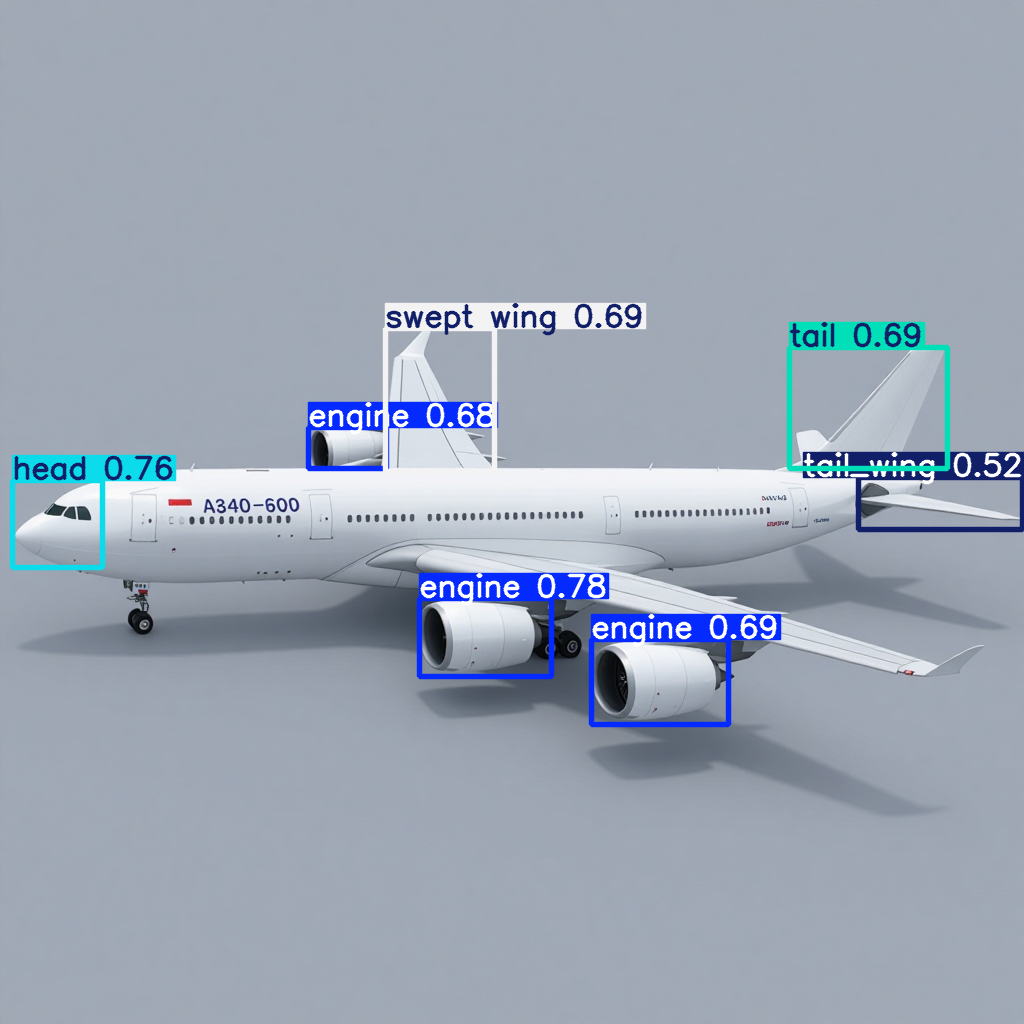}
        \caption{Image 3}
        \label{fig:img3}
    \end{subfigure}
    \hfill
    \begin{subfigure}[t]{0.23\textwidth}
        \centering
        \includegraphics[width=\linewidth]{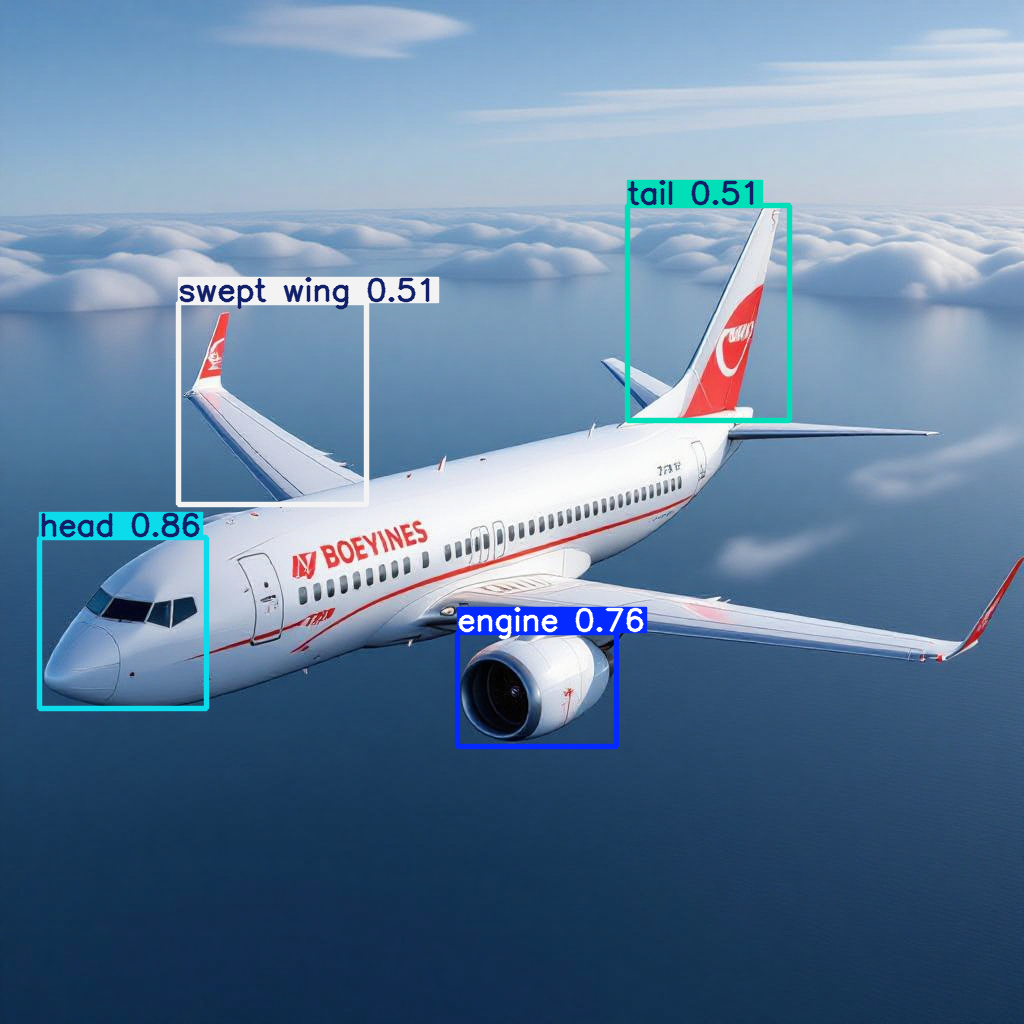}
        \caption{Image 4}
        \label{fig:img4}
    \end{subfigure}

    \caption{Sample aircraft images from the test dataset illustrating class variations.}
    \label{fig:eight_images_part1}
\end{figure}

\begin{figure}[H]
    \centering

    \begin{subfigure}[t]{0.23\textwidth}
        \centering
        \includegraphics[width=\linewidth]{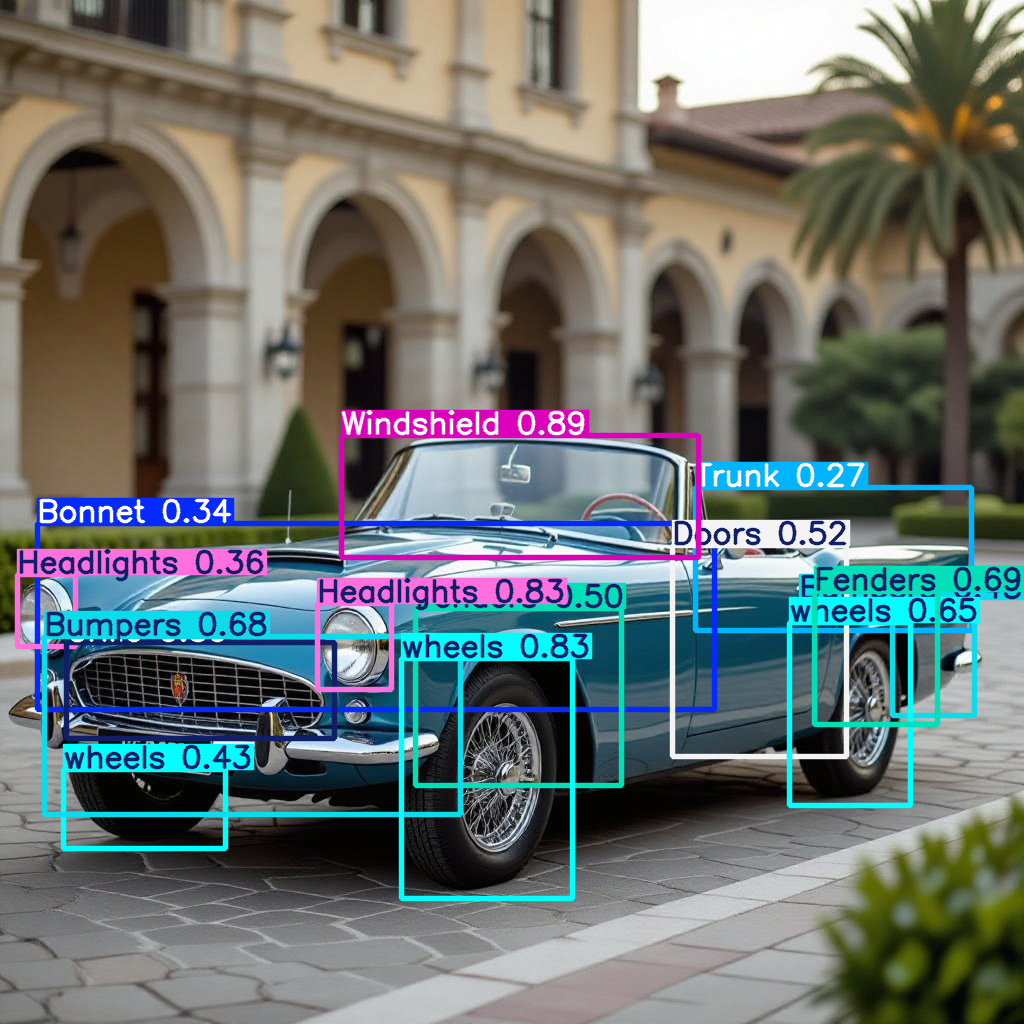}
        \caption{Image 5}
        \label{fig:img5}
    \end{subfigure}
    \hfill
    \begin{subfigure}[t]{0.23\textwidth}
        \centering
        \includegraphics[width=\linewidth]{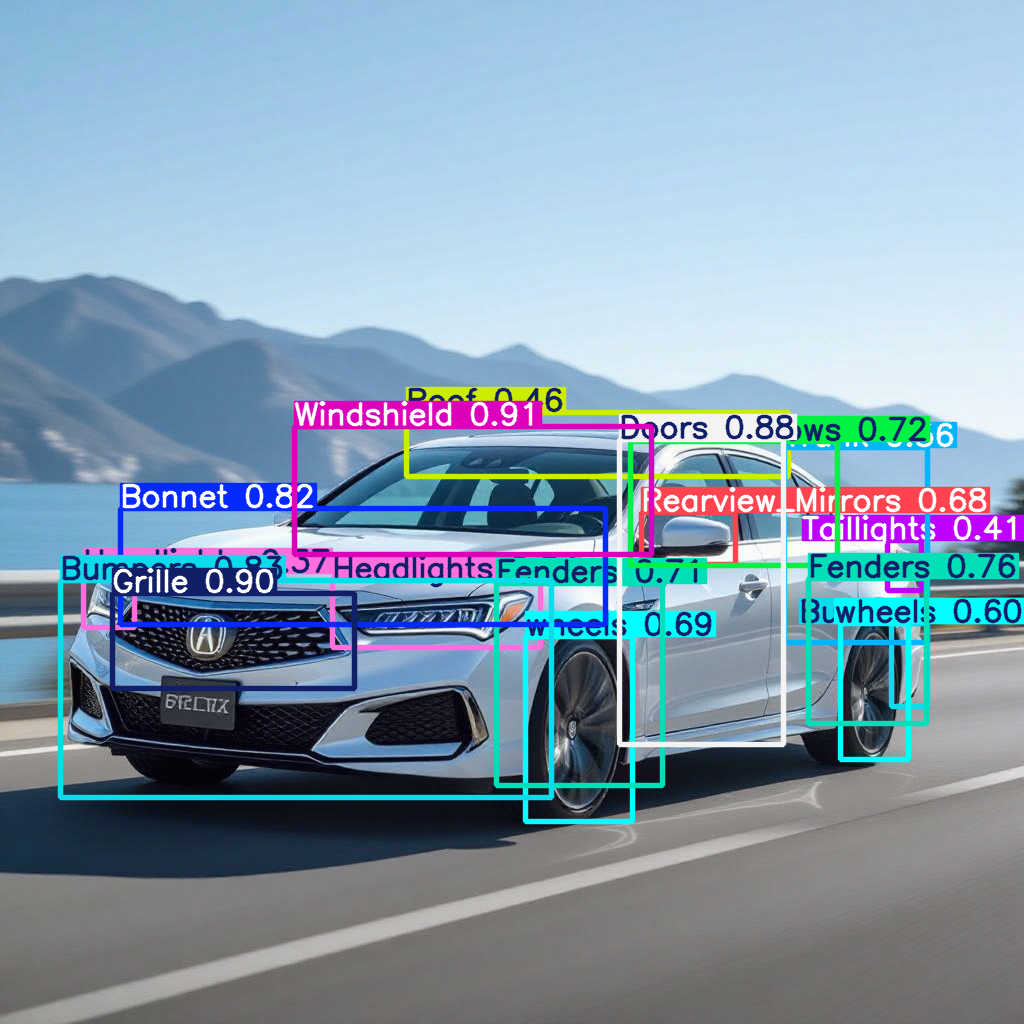}
        \caption{Image 6}
        \label{fig:img6}
    \end{subfigure}
    \hfill
    \begin{subfigure}[t]{0.23\textwidth}
        \centering
        \includegraphics[width=\linewidth]{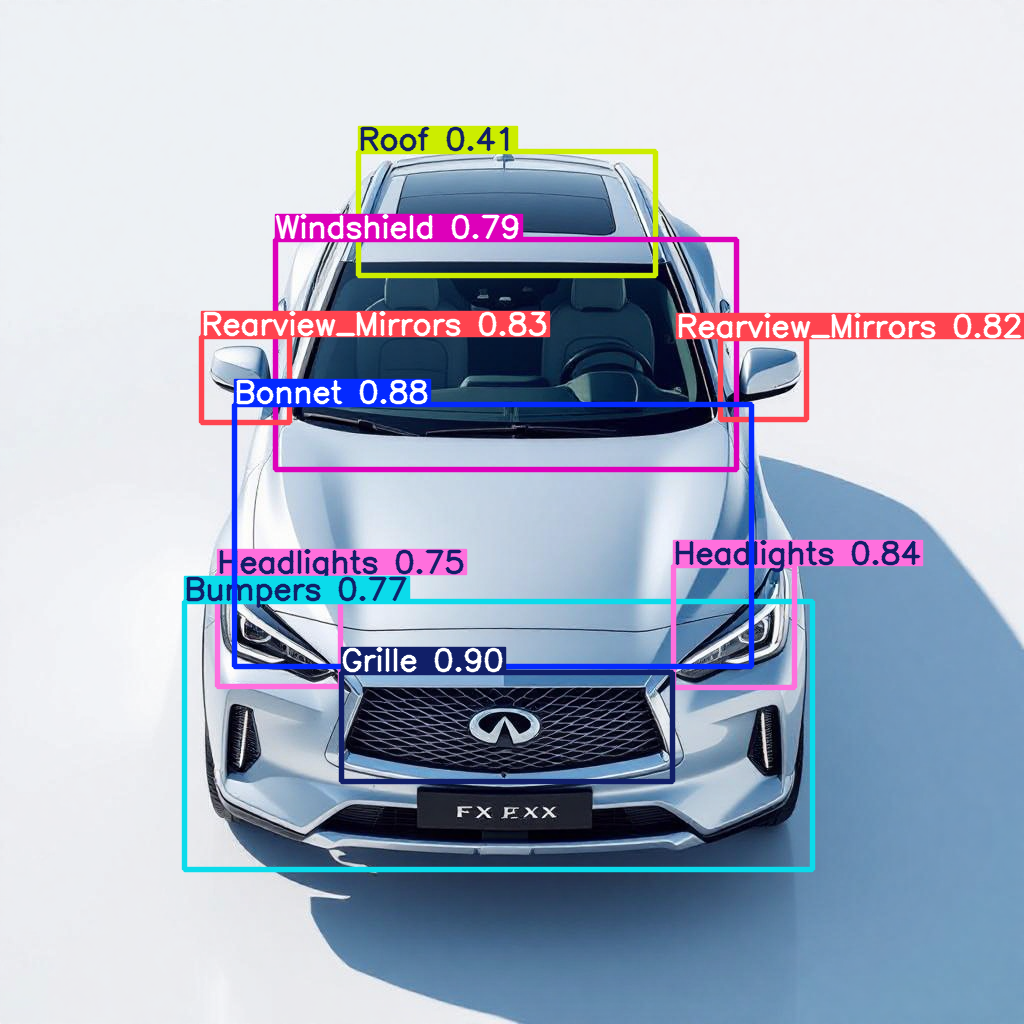}
        \caption{Image 7}
        \label{fig:img7}
    \end{subfigure}
    \hfill
    \begin{subfigure}[t]{0.23\textwidth}
        \centering
        \includegraphics[width=\linewidth]{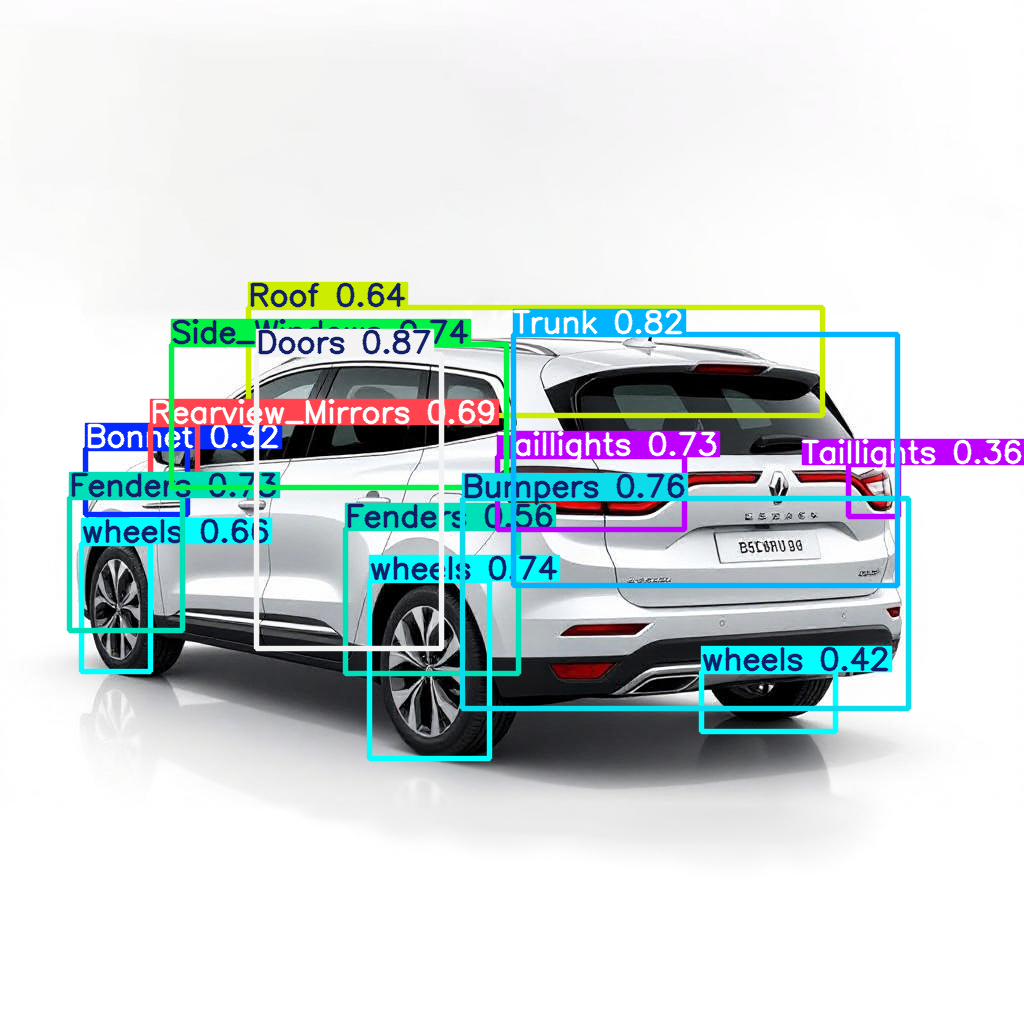}
        \caption{Image 8}
        \label{fig:img8}
    \end{subfigure}

    \caption{Sample car images from the test dataset illustrating class variations.}
    \label{fig:eight_images_part2}
\end{figure}

\subsubsection{Zero shot Detection Samples}

\begin{figure}[H]
    \centering

    \begin{subfigure}[t]{0.23\textwidth}
        \centering
        \includegraphics[width=\linewidth]{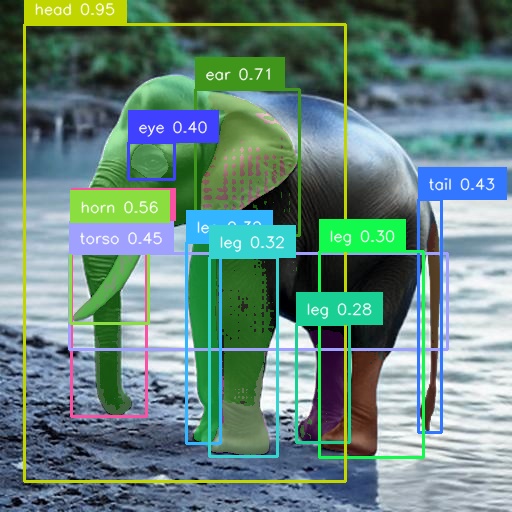}
        \caption{Image 5}
        \label{fig:img5_2}
    \end{subfigure}
    \hfill
    \begin{subfigure}[t]{0.23\textwidth}
        \centering
        \includegraphics[width=\linewidth]{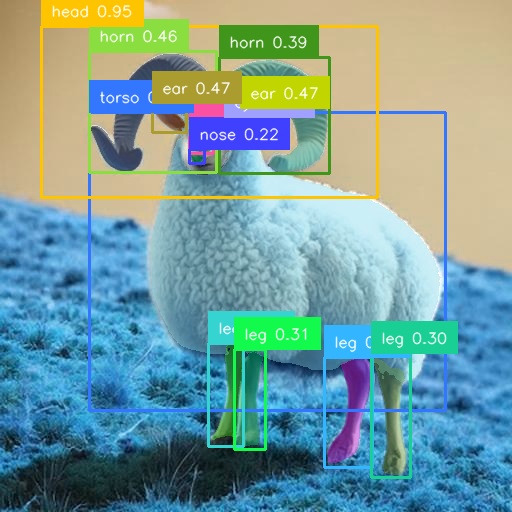}
        \caption{Image 6}
        \label{fig:img6_2}
    \end{subfigure}
    \hfill
    \begin{subfigure}[t]{0.23\textwidth}
        \centering
        \includegraphics[width=\linewidth]{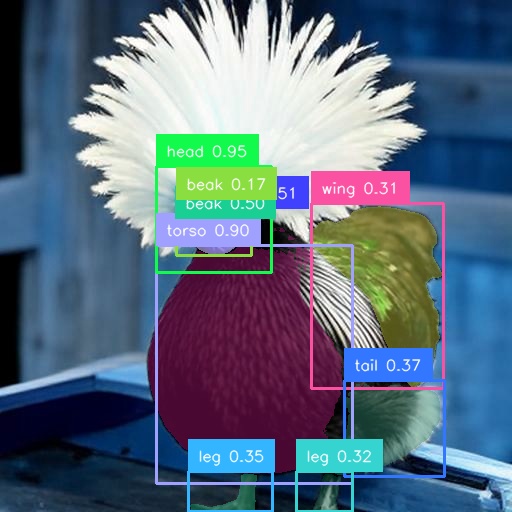}
        \caption{Image 7}
        \label{fig:img7_2}
    \end{subfigure}
    \hfill
    \begin{subfigure}[t]{0.23\textwidth}
        \centering
        \includegraphics[width=\linewidth]{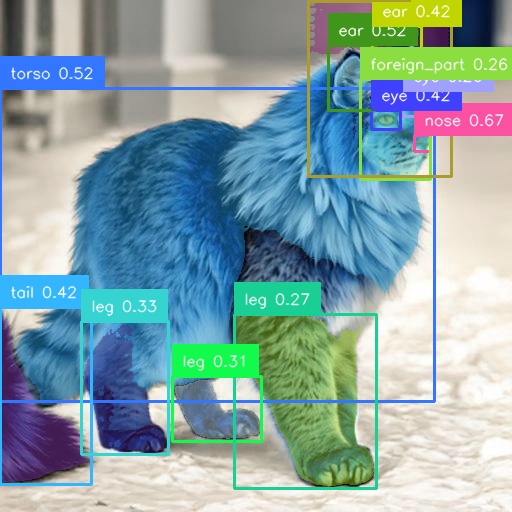}
        \caption{Image 8}
        \label{fig:img8_2}
    \end{subfigure}

    \caption{Sample animal images from the test dataset illustrating class variations.}
    \label{fig:eight_images_part2_1}
\end{figure}

We developed a two-stage vision pipeline, Grounding DINO and SAM, to accurately detect and segment animal body parts. First, Grounding DINO classifies the animal (or falls back to "mammal" or “bird”) to dynamically load specific body-part text prompts. It then detects these parts at multiple scales and generates initial bounding boxes for the head, torso, and limbs. We utilized specific Geometric Rules necessary for torso detection. To accurately identify the torso, we establish that if the head is detected, 50 percent of the total area of the identified animal will be encompassed within the torso region. If a leg is detected, then 50 percent of the area above the leg of the identified animal will correspond to the torso region. The refined boxes are subsequently transmitted to SAM, which generates precise, pixel-perfect masks for each distinct body part. Ultimately, we eliminate overlapping duplicates through Mask-based NMS utilizing precise pixel overlap and subsequently store the refined results along with JSON metrics.

\textbf{Geometric Rules.}
Let $B = (x_{\min}, y_{\min}, x_{\max}, y_{\max})$ denote a bounding box.
We apply three geometric rules for accurate detections prior to SAM segmentation:

\begin{itemize}
    \item \textbf{Head Reconstruction:} $B_{\text{Head}}$ is redefined to tightly
    bound all detected facial features (eyes, nose, etc.) plus a 20\% spatial padding.

    \item \textbf{Torso Rules:} To maintain the anatomical hierarchy, the vertical
    bounds of the torso are strictly cropped to start below the mid-head and end
    above the mid-legs.
    \begin{align}
        y_{\min}^{\text{torso}} &= \max\!\left(y_{\min}^{\text{torso}},\;
        y_{\text{mid}}^{\text{head}}\right) \\[4pt]
        y_{\max}^{\text{torso}} &= \min\!\left(y_{\max}^{\text{torso}},\;
        y_{\text{mid}}^{\text{legs}}\right)
    \end{align}

    \item \textbf{Limb Validity Filter:} To prevent background noises as detected
    as limbs or whole animal as a limb, $B_{\text{limb}}$ is discarded unless it
    satisfies these conditions:
    \begin{itemize}
        \item To prevent whole-body boxes:\\
        \textit{Size:} $\text{Area}(B_{\text{limb}}) \leq 1.2 \times
        \text{Area}(B_{\text{torso}})$
        \item Limbs must intersect the ``core body''
        $B_{\mathrm{limb}} \cap B^{\mathrm{pad}}_{\mathrm{core}} \neq \emptyset$ expanded by 15\% padding:\\
        \textit{Connectivity:} $B_{\text{limb}} \cap B_{\text{core}}^{\text{pad}}
        = \emptyset$
    \end{itemize}
\end{itemize}

\end{document}